\documentclass[twoside]{article}

\usepackage[accepted]{aistats2023}

\usepackage[round]{natbib}

\usepackage{mathtools}
\usepackage{amssymb}
\usepackage{amsmath}
\usepackage{amsthm}
\usepackage{graphicx} 
\usepackage[hidelinks]{hyperref}
\usepackage{xcolor}

\usepackage[ruled,vlined]{algorithm2e}

\makeatletter
\newtheorem*{rep@theorem}{\rep@title}
\newcommand{\newreptheorem}[2]{%
\newenvironment{rep#1}[1]{%
 \def\rep@title{#2 \ref{##1}}%
 \begin{rep@theorem}}%
 {\end{rep@theorem}}}
\makeatother

\newtheorem{theorem}{Theorem}
\newreptheorem{theorem}{Theorem}

\usepackage{subcaption}
\usepackage{caption}
\usepackage{float}
\usepackage{comment}
\usepackage{ulem}

\begin{document}

\twocolumn[

\aistatstitle{Efficient SAGE Estimation via Causal Structure Learning}

\aistatsauthor{ Christoph Luther* \And Gunnar K\"onig* \And  Moritz Grosse-Wentrup }

\aistatsaddress{University of Vienna\\ UniVie Doctoral School CS \And  LMU Munich \\ University of Vienna \\ Munich Center for ML (MCML) \And University of Vienna  \\ Data Science @ Uni Vienna \\ Vienna CogSciHub } ]




\begin{abstract}
The Shapley Additive Global Importance (SAGE) value is a theoretically appealing interpretability method that fairly attributes global importance to a model's features.
However, its exact calculation requires the computation of the feature's surplus performance contributions over an exponential number of feature sets. 
This is computationally expensive, particularly because estimating the surplus contributions requires sampling from conditional distributions. 
Thus, SAGE approximation algorithms only take a fraction of the feature sets into account. 
We propose $d$-SAGE, a method that accelerates SAGE approximation. 
$d$-SAGE is motivated by the observation that conditional independencies (CIs) between a feature and the model target imply zero surplus contributions, such that their computation can be skipped.
To identify CIs, we leverage causal structure learning (CSL) to infer a graph that encodes (conditional) independencies in the data as $d$-separations. 
This is computationally more efficient because the expense of the one-time graph inference and the $d$-separation queries is negligible compared to the expense of surplus contribution evaluations. Empirically we demonstrate that $d$-SAGE enables the efficient and accurate estimation of SAGE values. 
\end{abstract}

\section{INTRODUCTION \label{section-intro}}

Machine learning (ML) is increasingly deployed in various fields, ranging from the sciences \citep{reichstein2019deep,schmidt2019recent,luan2021review,farrell2018novel} to high-stakes decisions about individuals \citep{Manish2020,zeng2015interpretable,obermeyer2019dissecting}. 
Despite impressive successes in predictive performance \citep{senior2020improved,bhatt2020explainable}, the complexity of ML models makes it difficult to assess their trustworthiness or to gain knowledge about the data generating process.
In recent years, the advent of interpretable machine learning has brought about a plethora of methods that provide insight into model and data \citep{molnar2020interpretable}. 
Among those, interpretability methods based on the Shapley value from game theory \citep{shapley1953stochastic} have gained popularity as they satisfy desirable fairness properties \citep{vstrumbelj2014explaining,datta2016algorithmic,lundberg2017unified,sundararajan2020many,covert2020understanding}.

SAGE values \citep{covert2020understanding} apply Shapley values to fairly attribute the model's predictive performance to the features, thereby providing valuable insight into dependencies in the data.
They are particularly appealing for scientific inference since they can be linked to properties of the data generating process \citep{covert2020understanding,freiesleben2022scientific}.
The building blocks for SAGE values are so-called SAGE value functions $\nu(\textbf{X}_S)$ that measure the performance contribution of arbitrary subsets of features $\textbf{X}_S$.
Based on these value functions, a feature's importance value $\phi$ is computed as the average surplus contribution $\nu(\textbf{X}_{S \cup j}) - \nu(\textbf{X}_S)$ of the feature $X_j$ over all possible subsets $\textbf{X}_S$ of the remaining features. 
This is a computationally demanding procedure due to the number of coalitions $\textbf{X}_S$ that grows exponentially with the number of features \citep{covert2020understanding,van2022tractability} and the high expense  of evaluating $\nu$ which stems from the conditional sampling that is required for its estimation. In practice, \citep{covert2020understanding} address the exponential number of coalitions by only computing the respective surplus contribution for a randomly sampled subset of the coalitions.\footnote{Furthermore, \cite{covert2020understanding} avoid conditional sampling for the evaluation of $\nu$ by employing marginal sampling instead. If features are dependent, this leads to extrapolation and does not allow linking the SAGE values to properties of the data generating process \citep{chen2020true}. In this work, we focus on estimating conditional SAGE values.}

In this work, we suggest exploiting the dependence structure in the data to speed up the estimation of (conditional sampling based) SAGE values in an approach we coin $d$-SAGE. 
More specifically, we show that the surplus contribution $\nu(\textbf{X}_{S \cup j}) - \nu(\textbf{X}_S)$ is zero for optimal predictors if the variable of interest is conditionally independent of the model’s target given the respective subset of remaining features (i.e., if $X_j \perp Y | \textbf{X}_S$, Theorem \ref{theorem:surplus-zero}).
As such, if we know the conditional independencies (CIs) in the data, the respective value function evaluations can be skipped. Since, in general, the dependence structure is unknown, and conditional independence testing is expensive, we leverage research in causal structure learning (CSL) that allows us to greedily learn graphical models which encode the dependence structure in the data. 

Overall, the approach is based on the following rationale: The quality of SAGE approximation hinges on the number of evaluations of $\nu$ that each require estimating conditional expectations and thus are computationally expensive.\footnote{The expense of the computation depends on the type of data for which the conditional expectation shall be computed. Previous work in the field assumes polynomial complexity for the operation \citep{van2022tractability}.} $d$-SAGE relies on the one-time estimation of a causal graph, which in practice can be performed by greedy-search algorithms in polynomial time \citep{scutari2019greedy}. The estimated graph then allows to identify CIs using linear-time $d$-separation queries \citep{hagberg2008networkx, darwiche2009modeling}. Every found $d$-separation, in turn, warrants to spare an expensive evaluation of $\nu(\textbf{X}_{S \cup j}) - \nu(\textbf{X}_S)$. Since graph learning has to be performed only once and $d$-separation queries are highly efficient, the runtime of SAGE estimation can be reduced significantly by skipping the computation of $\nu(\textbf{X}_{S \cup j}) - \nu(\textbf{X}_S)$ whenever warranted. 
We show empirically that the saved runtime is approximately equal to the share of CIs. 

\subsection{Contributions \label{subsection-contributions}}
We propose $d$-SAGE, the first method that exploits the dependence structure in the data to make SAGE estimation more efficient. More specifically, we find that CIs in the data imply that the respective (expensive) surplus evaluations can be skipped and suggest leveraging greedy CSL for their identification (Section \ref{section:csl-for-sage}).
To select a suitable CSL algorithm, we perform a benchmark that, in contrast to previous work, evaluates the algorithms' ability to efficiently identify CIs in the data (Section \ref{section:csl-benchmark}).
On twelve synthetic datasets, we demonstrate empirically that $d$-SAGE and the approximation algorithm by \cite{covert2020understanding} converge towards the same estimates but that $d$-SAGE is significantly faster. We find that the computational overhead of learning the causal structure is negligible compared to the computational cost of the surplus evaluations, such that the overall runtime reduction is approximately equal to the share of CIs found in the data (Section \ref{section:sage-experiments}).
Consequently, $d$-SAGE enables the application of SAGE for larger models, especially in sparse settings.

\section{RELATED WORK \label{section-relatedwork}}

While there are many attempts to tackle the complexity of Shapley value based methods,
most existing work targets speeding up SHAP \citep{lundberg2017unified} estimation \citep{jethani2021fastshap, covert2021improving, li2020efficient} or is limited to be applied with random forests \citep{benard2022shaff}.
In contrast, our work is model-agnostic and targets improving SAGE estimation.
Moreover, none of the existing work exploits the dependence structure in the data to yield efficiency gains. As such, we see our work as complementary to the approach of \cite{mitchell2022sampling}, who suggest to carefully select permutations. 

In recent years, concepts from causality have also been introduced to Shapley value based importance measures to adapt them to answer specific questions or to improve model interpretation. \cite{asv}, for example, introduce \textit{asymmetric Shapley values} that can either shift the explanatory power of all variables along a causal chain towards the root cause (distal approach) or towards immediate causes (proximate approach). Moreover, \cite{causalsvs} use Pearl's do-calculus to develop \textit{causal Shapley values} and \cite{wang2021shapley} propose to attach importance to edges in a causal graph instead of explanatory variables, i.e., nodes in the graph.
In contrast to the literature, we seek efficiency gains for feature attributions from causal inference research while retaining the principle of SAGE values unaltered.

We do, however, make use of CSL. \cite{scutari2019betterbn} and \cite{const2021large} provide large-scale benchmark studies of structure learning algorithms. In short, both studies agree on the superiority of score-based structure learning based on greedy search algorithms over constraint-based and hybrid methods. These findings motivate our choice of CSL algorithms for $d$-separation inference. In contrast to existing work, our benchmark does not focus on recovering the causal structure but on detecting CIs in the data.

\section{BACKGROUND \label{section-background}}

This section serves to familiarise the reader with the basic concepts required to understand this paper. First, we introduce SAGE values for global feature importance (Section \ref{subsection-sage}). Then, we recapitulate why SAGE values are difficult to estimate and present the SAGE approximation algorithm (Section \ref{subsection-tractability}). Last we explain CSL, which we later use to speed up SAGE estimation (Section \ref{subsection-csl}).

\subsection{Shapley Additive Global Importance \label{subsection-sage}}

The Shapley value, which was initially proposed in game theory \citep{shapley1953stochastic}, is commonly applied for feature relevance quantification \citep{vstrumbelj2014explaining,datta2016algorithmic,lundberg2017unified,sundararajan2020many,covert2020understanding}. In the study of cooperative games, it serves to fairly attribute the outcome of a game to all participating players. The principle can be applied to assess the relevance of variables for a predictor $f$, where the predictive performance is the outcome of the game and the variables are the players. 
\cite{covert2020understanding} leverage Shapley values to derive a \textit{global} measure of feature importance, i.e. SAGE values. Global in this context means that the importance of a feature across all instances in a sample is assessed. For an arbitrary model $\hat{f}$ using inputs $x_1,...,x_d$, \cite{covert2020understanding} define the SAGE value for the $j$-th feature as:

\begin{equation} \label{eq:sv} 
    \begin{aligned}
        \phi_j(\nu) =  \frac{1}{d!}  \sum_{\pi \in \Pi(d)} & \big(\nu(\{X_i: \pi(i) \leq \pi(j) \}) \\ &- \nu(\{X_i: \pi(i) < \pi(j) \})\big) 
    \end{aligned}
\end{equation}

where $X_j$ is the random variable corresponding to feature observation $x_j$, $\Pi(d)$ is the set of all permutations of indices $\{1,...,d\}$, $\pi$ a specific permutation and $\pi(j)$ the position of feature $j$ in permutation $\pi$. For the sake of readability, we use the more general notation $\textbf{X}_S$ instead of $\{X_i: \pi(i) < \pi(j) \}$ as input to the value function $\nu$ with $\textbf{X}_S$ being any set of features and $S$ the collection of indices of the contained features, i.e. $S \subseteq \{1,...,d\}$ ($\bar{S}$ is its complementary set). $\nu(\textbf{X}_S)$ is defined as
\begin{equation*}
 \nu(\textbf{X}_S) = \mathbb{E}_{\textbf{X}, Y}[\ell(\hat{f}_{\emptyset}(\textbf{X}_{\emptyset}),Y)] - \mathbb{E}_{\textbf{X}, Y}[\ell(\hat{f}_{S}(\textbf{X}_S),Y)],
\end{equation*}
where $\ell(\cdot)$ is any admissible loss function and $\hat{f}_S(\textbf{x}_S) = \mathbb{E}_{\textbf{X}_{\bar{S}}|\textbf{X}_S}[\hat{f}(\textbf{X})|\textbf{X}_S=\textbf{x}_S]$. Thus, $\nu(\textbf{X}_S)$ is the reduction in risk induced by adding $\textbf{X}_S$. Consequently, SAGE values gauge a feature $j$'s importance using the average over the additional reduction in risk of the feature compared to any existing coalition.

SAGE values are particularly appealing as they satisfy six desirable fairness axioms that set them apart from other feature importance measures: \textit{efficiency}, the \textit{dummy property}, \textit{symmetry}, \textit{monotonicity}, \textit{linearity}\footnote{For simplicity we employ the names of these Shapley value properties for the SAGE properties that are described in Appendix \ref{appendix-sage}.} and invariance to monotone transformations.
Despite a thorough mathematical foundation and the fulfilment of mentioned desiderata, SAGE values have a major drawback: They require the evaluation of an exponential number of surplus evaluations, which is computationally infeasible. In practice, only a subset of possible coalitions is evaluated (cf. Section \ref{subsection-tractability}). 

To estimate SAGE values, access to the conditional feature distributions is required;
More specifically, we need to sample from $P(\textbf{X}_{\bar{S}}|\textbf{X}_S)$ to estimate the marginalized prediction $\hat{f}_S(\textbf{x}_S) = \mathbb{E}_{\textbf{X}_{\bar{S}}|\textbf{X}_S}[\hat{f}(\textbf{X})|\textbf{X}_S = \textbf{x}_S]$.  
Problematically, conditional samplers may not be readily available in practice. \cite{covert2020understanding} suggest eluding the problem by sampling from $P(\textbf{X}_{\bar{S}})$ instead (marginal sampling).
Albeit easy to implement (and computationally efficient), marginal sampling may generate unrealistic data points $(\textbf{x}_S, \textbf{x}_{\bar{S}})$ and thus marginal-sampling based SAGE values are not suitable for inference about the data generating process or to understand the model's behaviour in the observational distribution \citep{frye2020manifold,chen2020true,aas2021explaining,molnar2022general}. Therefore, we focus on conditional SAGE and estimate the conditional distributions if they are not known a-priori.

To estimate conditional distributions a variety of techniques can be employed: For categorical variables, estimating the conditional reduces to standard supervised learning with cross-entropy loss. For linear Gaussian data, the conditional can be estimated analytically from the covariance matrix \citep{page1984multivariate}. A range of methods exist for continuous settings with nonlinearities \citep{bishop1994mixture,bashtannyk2001bandwidth,sohn2015learning,trippe2018conditional,winkler2019learning,hothorn2021predictive}. For mixed data, a sequential design can be used \citep{blesch2022conditional}.\\

\subsection{Intractability of SAGE and Approximation Algorithm \label{subsection-tractability}}

For the Shapley based interpretability approach SHAP intractability was proven \citep{van2022tractability}. For the exact computation, the surplus contribution for all possible subsets of the remaining features must be evaluated. The number of possible subsets grows exponentially in the number of features.

Exact SAGE estimation also suffers from the exponential number of coalitions. To address the issue, \cite{covert2020understanding} propose an approximation algorithm that does not take all possible coalitions into account. More specifically, the authors propose to repetitively sample permutations $\pi$ from the feature indices. Then, for every element of the current permutation, starting with the first one, they successively compute $\Delta_{j|S} := \nu(\textbf{X}_{S \cup j}) - \nu(\textbf{X}_S)$ with the set $S$ being all features that come before the feature of interest $j$ in $\pi$. 
We yield the estimated importance $\hat{\phi}_j(\nu)$ for $X_j$ by taking the mean of all $\Delta_{j|S}$ values over the different permutations $\pi$.
The approximation algorithm is unbiased and the variance of the estimate reduces in $O(\frac{1}{n})$ \citep{covert2020understanding}. However, considering the risk evaluation required for estimating $\nu$, the procedure based on conditional sampling remains computationally demanding.

\subsection{Causal Structure Learning \label{subsection-csl}}

This section deals with the introduction of CSL used to estimate graphs representing $d$-separations. $d$-separation is the graphical equivalent to conditional independence in the underlying distribution. Both concepts are indeed equivalent under two standard assumptions: (1) that the Markov property is fulfilled and (2) that the distribution is faithful w.r.t. the graph. We write $X_j \perp_{\mathcal{G}} Y | \textbf{X}_S$ when a variable $X_j$ is $d$-separated from $Y$ given $\textbf{X}_S$ in a graph $\mathcal{G}$. 
Since we merely use graphs to read off $d$-separations, we leave out a holistic coverage and refer the reader to \citet{darwiche2009modeling} and \citet{pearl_2009}. Here, it shall suffice that we refer to a directed acyclic graph (DAG) whose nodes represent random variables from the underlying distribution and  whose edges reflect direct dependencies in the data. Edge directions are further interpreted as cause-effect relations. We now briefly summarise the inference of such graphs from data.

Generally, one distinguishes between \textit{constraint-based} and \textit{score-based} methods. The former use CIs inferred from data as constraints on where to draw edges. The latter explore the space of all possible DAGs over the given variables and assign scores to every visited graph. The output of the algorithm is the highest scoring graph. Since the space of DAGs over a set of variables or nodes grows superexponentially in the set's cardinality, score-based methods often rely on greedy search techniques. In addition, \textit{hybrid methods} combine both CIs as constraints and scoring of graphs to assess candidates.

In this work, we focus on greedy structure learning that performed best in recent benchmarks \citep{scutari2019betterbn, const2021large}. More precisely, we rely on structure inference based on hill-climbing (HC) and TABU search \citep{russel2010, scutari2019greedy}. Crucially, both algorithms use the Bayesian information criterion \citep{schwarz1978bic}, which satisfies two key properties, \textit{consistency} and \textit{local consistency}\footnote{The Bayesian Dirichlet equivalent uniform (BDeu) score satisfies the properties too and is a valid alternative.} \citep{gamez2011hc, chickering2002greedy}. \cite{gamez2011hc} show that for HC for a dataset of size $n$ and \textit{iid} data, the output graph is a minimal I-Map of the underlying distribution if $n \rightarrow \infty$ and the scoring function satisfies \textit{consistency} and \textit{local consistency}. By definition of a minimal I-Map, the set of CIs represented by $d$-separation in the graph is a subset of the CIs in the distribution. Hence, while there might be independencies in the underlying distribution of the data not represented by $d$-separation, there are no instances of $d$-separations that do not correspond to independencies. Note that HC introduces a DAG structure of the output graph but the assumption on the data is just being an \textit{iid} sample. The proof, however, hinges on the assumption of faithfulness. For linear models, though, the probability of faithfulness being violated is shown to be zero if model parameters are randomly drawn from positive densities (cf. \cite{Peters2017}, \cite{spirtes2000causation}). While there is no similar theoretical result for TABU, the latter is an extension of HC and exhibits similar behaviour in practice (cf. Section \ref{section:csl-benchmark}).


\section{CAUSAL STRUCTURE LEARNING FOR EFFICIENT SAGE ESTIMATION \label{section:csl-for-sage}}

SAGE estimation is computationally challenging. For an exact computation, the surplus contribution of the feature of interest $j$ with respect to every possible coalition $\textbf{X}_S$ of the remaining features must be computed. The surplus contribution is defined as in Section \ref{subsection-tractability}
\begin{equation}\label{eq:delta}
\Delta_{j|S} = \nu(\textbf{X}_{S \cup j}) - \nu(\textbf{X}_S)
\end{equation}
The number of possible coalitions grows exponentially in the number of features, making the exact computation intractable in high-dimensional settings.
SAGE values are therefore estimated by randomly sampling coalitions until the estimates converge (Section \ref{subsection-tractability}). Nevertheless, estimation remains challenging since evaluating $\Delta_{j|S}$ requires sampling from conditional distributions, and therefore even one evaluation is a significant computational challenge. Thus, in practice, the approximation quality is limited by the number of surplus contributions that can be computed.

We propose $d$-SAGE, an approach that can identify and skip unnecessary surplus evaluations and thereby allows to improve the approximation quality.
The method is based on the observation that $\Delta_{j|S}$ evaluates to zero if $X_j$ is conditionally independent of $Y$ given $X_S$: 

\begin{theorem}

For $\ell$ being cross-entropy loss or the mean-squared error, $f^*$ the respective optimal predictor and $\nu_{\ell,f^*}$ the corresponding SAGE value function, it holds that
\begin{align*}
    X_j \perp Y | \textbf{X}_S \Rightarrow \nu_{\ell,f^*}(\textbf{X}_{S \cup j}) - \nu_{\ell,f^*}(\textbf{X}_S) = 0.
\end{align*}

\label{theorem:surplus-zero}
\end{theorem}
\textit{Proof (sketch, full proof in \ref{sec:theorem:surplus-zero}): \cite{covert2020understanding} show that for the cross entropy loss function with its respective optimal model, the Bayes classifier, Equation \ref{eq:delta} equals the conditional mutual information of $X_j$ and $Y$ given $\textbf{X}_S$, i.e. $I(X_j;Y|\textbf{X}_S)$. A similar result holds for optimal regression models with the mean squared error (MSE) as loss function. In this case, the surplus contribution is shown to be equal to $\mathbb{E}_{\textbf{X}_S}[Var(\mathbb{E}[Y|\textbf{X}_S, X_j]|\textbf{X}_S)]$ \citep{covert2020understanding}. For both expressions, one can easily see that they evaluate to zero when $X_j$ is conditionally independent of $Y$ given $\textbf{X}_S$, i.e. when $X_j \hspace{0.1cm} \perp \hspace{0.1cm} Y \hspace{0.1cm} | \hspace{0.1cm} \textbf{X}_S$.}

As a consequence of Theorem \ref{theorem:surplus-zero}, knowledge of the dependence structure in the data allows speeding up the SAGE estimation procedure: evaluations of $\nu(\textbf{X}_{S \cup j}) - \nu(\textbf{X}_S)$ can be skipped if $X_j \hspace{0.1cm} \perp \hspace{0.1cm} Y \hspace{0.1cm} | \hspace{0.1cm} \textbf{X}_S$.


To identify the CIs in the data, we suggest leveraging greedy procedures that were originally developed to learn the causal structure in the data. CSL algorithms allow the estimation of a causal graph in polynomial time \citep{scutari2019greedy}. Given that the Markov property and faithfulness are fulfilled, the graph allows reading off (conditional) independencies in the data using linear time $d$-separation queries \citep{hagberg2008networkx, darwiche2009modeling}. Our rationale is that the one-time effort of learning the causal graph, as well as the additional linear time $d$-separation queries, are negligible in comparison to the computational overhead of computing the surplus contributions.\footnote{In general, the complexity of conditional sampling depends on the assumptions about the data generating process. In their tractability analysis for SHAP, \cite{van2022tractability} assume polynomial complexity for computing the conditional expectations of the form $\mathbb{E}_{\textbf{X}_{\bar{S}}|\textbf{X}_S}[\hat{f}(\textbf{X})|\textbf{X}_S=\textbf{x}_S]$.}

To summarise, $d$-SAGE estimation introduces two key differences to the original SAGE approximation algorithm. First, a graph $\mathcal{G}$ is fitted over all random variables, the features, and the target. Second, the estimation of $\Delta_{j|S}$ is skipped if the current feature $X_j$ in permutation $\pi$ is d-separated from the target given the set $\textbf{X}_S = \{ X_i : \pi(i) < \pi(j) \}$. The changes are highlighted in blue in Algorithm \ref{alg:dsage}.

\normalem 
\begin{algorithm}[t]  
    \SetAlgoLined 
    \textbf{Input:} Data $\{\textbf{x}_i, y_i \}_{i=1}^n$ with $\textbf{x}_i \in \mathbb{R}^d$, model $\hat{f}$, loss function $\ell$, number of permutations $n_{\pi}$  \\
    \textcolor{blue}{Infer \textbf{DAG} $\mathcal{G}$ from data $\{\textbf{x}_i, y_i \}_{i=1}^n$ with structure learning algorithm of choice.} \\
    \For{i in $\{1,...,n_{\pi}\}$}{
        Sample a permutation $\pi$ \\
        $S = \emptyset$ \\
        \For{j in $\{1,...,d\}$}{
            \eIf{\textcolor{blue}{$X_{\pi_j} \not\perp_{\mathcal{G}} Y | \textbf{X}_S$}}{
            
                Sample $\textbf{x}_{\bar{S}}$ from $p(\textbf{x}_{\bar{S}}|\textbf{x}_S)$ \\
                
                Sample $\textbf{x}_{\overline{S \cup \pi_j}}$ from $p(\textbf{x}_{\overline{S \cup \pi_j}}|\textbf{x}_S)$, where $\pi_j$ is the $j$-th element of $\pi$ \\
                
                $\hat{\Delta}_{j|S} = \ell( \hat{f}(\textbf{x}_S, \textbf{x}_{\bar{S}}) ) - \ell (\hat{f} (\textbf{x}_{S \cup \pi_j }, \textbf{x}_{\overline{S \cup \pi_j}}) )$
        }{ \textcolor{blue}{$\hat{\Delta}_{j|S} = 0$}}
    
        $S = S \cup \pi_j$
        }
    }

    \textbf{return} $\hat{\phi}_j = \frac{1}{n_{\pi}} \sum_{i = 1}^{n_{\pi}} \hat{\Delta}_{j|S}$  \hspace{0.05cm} for $j=1,...,d$  \\[0.3cm]

    \caption{Sampling-based Approximation of $d$-SAGE \label{alg:dsage}}

    \footnotesize Note that we dropped indices of $\hat{\Delta}_{j|S}$ for readability.
\end{algorithm} 
\ULforem 

\section{EXPERIMENTS \label{section-experiments}} 

This section is divided into three parts.
In the first two parts, we evaluate our method on synthetic data with known ground truth: As we use $d$-separation queries in estimated graphs for $d$-SAGE approximation, we first evaluate the accuracy of $d$-separations in learned structures with regard to ground truth CIs in the data (Section \ref{section:csl-benchmark}). Then we compare $d$-SAGE 
to ordinary SAGE value approximation (Section \ref{section:sage-experiments}). 
In the third part, we demonstrate the usefulness of the method in a real-world application (Section \ref{subsec:experiments:real-world-application}).\footnote{All code is publicly available \url{https://github.com/gcskoenig/csl-experiments/tree/camera-ready}.}\\

\subsection{Benchmark of Causal Structure Learning \label{section:csl-benchmark}}

Existing structure learning benchmarks evaluate the algorithms regarding how well they can recover the true causal structure 
\citep{const2021large, scutari2019betterbn}. 
For $d$-SAGE, however, we are only interested in learning the dependence structure. As such, 
we assess how well CIs in the data are represented as $d$-separations in the estimated graph. 

\subsubsection{Setup}

We evaluate the greedy search algorithms HC and TABU \citep{scutari2019greedy,russel2010}.
We selected these methods based on their superior performance in recent CSL benchmarks \citep{const2021large,scutari2019betterbn}.
As performance metrics, we employ the F1 score for the detection of $d$-separations w.r.t. a randomly sampled target $Y$ as well as the respective false discovery rate.
More precisely, for every potential $d$-separation of the form $X_j \perp_{\mathcal{G}} Y | \textbf{X}_S$ 
, we check whether it had the same status in the ground truth and the estimated graph. To cope with the exponentially large number of $d$-separations in the higher dimensional graphs (DAG$_{sm}$, DAG$_{m}$ and DAG$_{l}$) we randomly sampled a node of interest $X_j$ and a conditioning set \textbf{X}$_S$ one million times instead of iterating over all potential $d$-separation statements. For both algorithms, we relied on their implementation in \textit{bnlearn} \citep{scutari2010bnlearn} for R.\footnote{All graph learning experiments were run on an Intel Core i7-8700K Desktop CPU.} We consider twelve different synthetic data settings with known ground truth:

\paragraph{\textbf{DAG$_s$}, \textbf{DAG$_{sm}$}, \textbf{DAG$_m$} and \textbf{DAG$_l$}}{
We sampled synthetic graphs with a varying number of nodes ($s=10$, $sm=20$, $m=50$ and $l=100$) and 
three different densities (average adjacency degrees of $2$, $3$ and $4$). Based on the graphs, we sampled data from the corresponding linear Gaussian data model, 
where absolute values of edge weights are bounded by $0.5$ and $2$.
We standardised variances to be (approximately) one to avoid that they increase with the topological ordering and counteract a potential bias in the benchmark
\citep{reisach2021beware}. 
For the sampling itself, we relied on the 
the \textit{pcalg} package \citep{kalisch2012pcalg} implemented in R \citep{rcore2022}.}









\subsubsection{Results}

\begin{figure}[ht]
\includegraphics[scale=0.5]{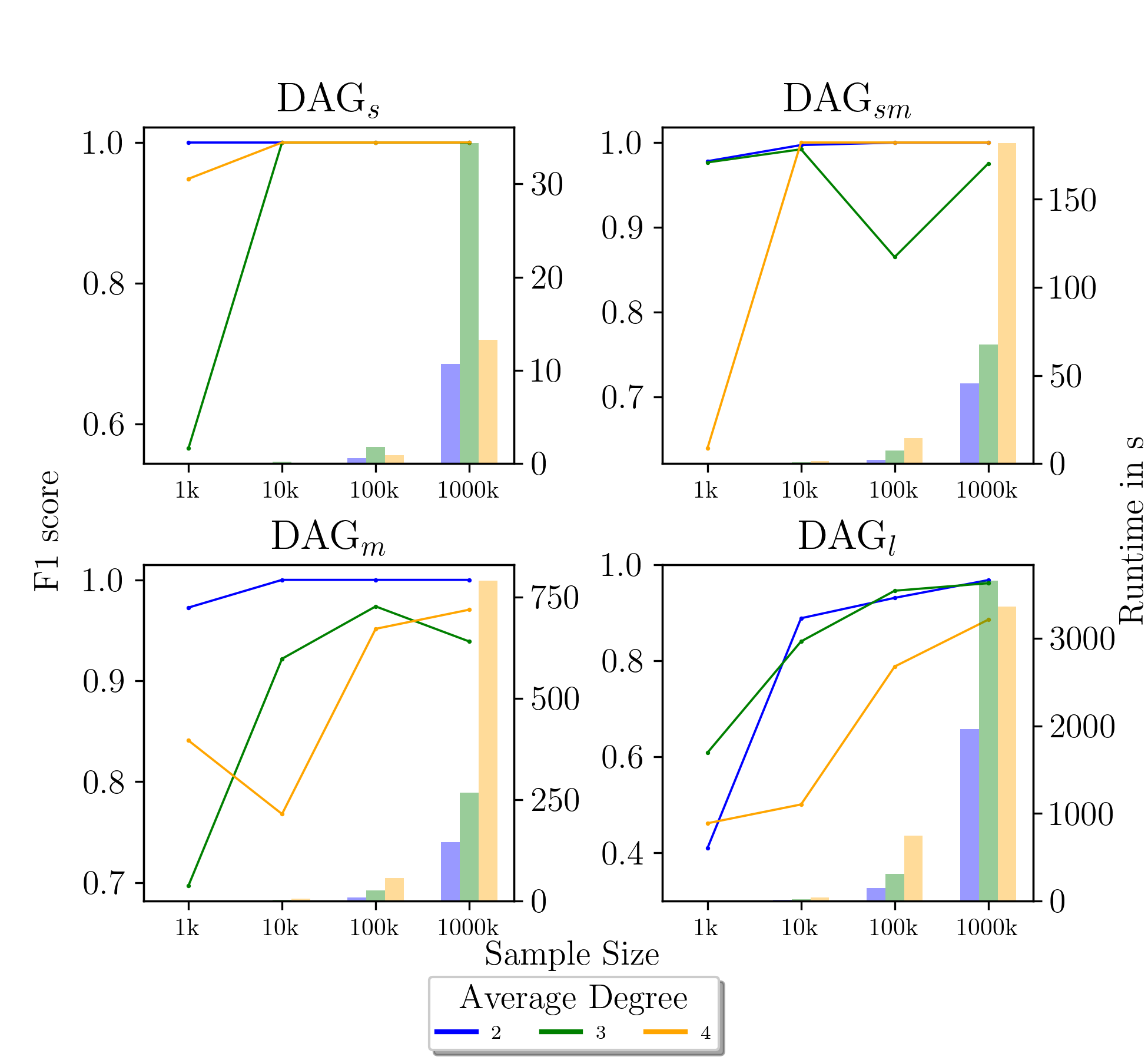}
\caption{F1 scores for $d$-separation (lines, left y-axes) and runtime of graph learning (bars, right y-axes) using TABU search depending on sample size. \label{fig:cslruntime}}
\end{figure}

\begin{figure}[ht]
\includegraphics[scale=0.55]{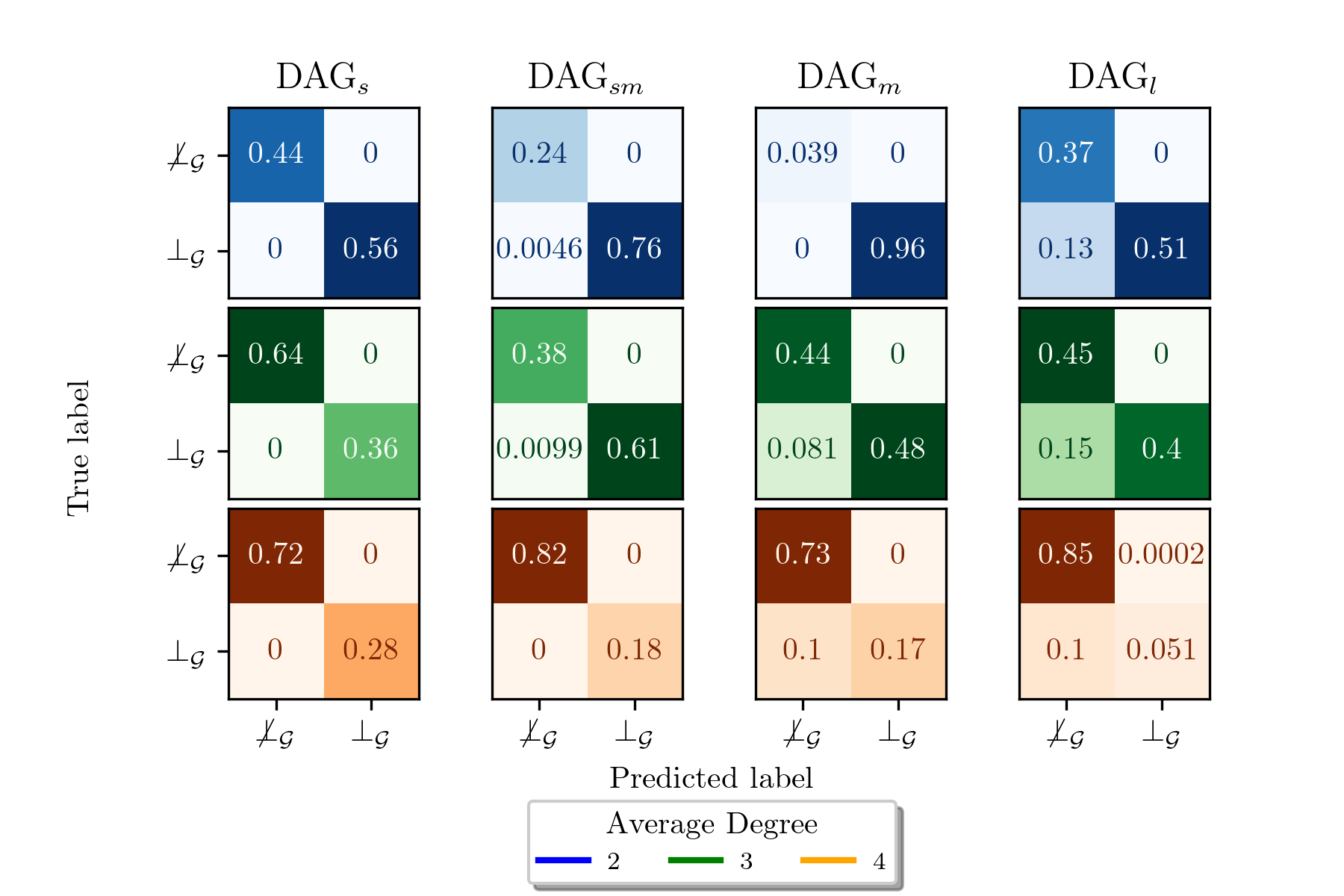}
\caption{Confusion matrix for true and predicted $d$-connections ($\not \perp_{\mathcal{G}}$) and $d$-separations ($\perp_{\mathcal{G}}$) based on TABU search with $n=10,000$  for all twelve graphs. \label{fig:confusion}}
\end{figure}

First, we observe that TABU, while approximately taking double the time, either performs equally well as or better than HC (cf. Figures \ref{fig:cslruntime}, \ref{fig:confusion} and Appendix \ref{appendix-graphs}). Hence, we restrict this section to results for TABU search, which we also employed for $d$-SAGE estimation. Figure \ref{fig:cslruntime} shows the runtime of graph learning depending on sample size and corresponding F1 scores for $d$-separation inference for all twelve graphs. The key takeaway is that for the sparsest graph (average adjacency degree $2$) the F1 score is greater than $0.88$ if $n \geq 10,000$. For the larger graphs, however, there is a slight drop-off in performance, which is expected. Only for the densest graph setting (average adjacency degree $4$) and for $50$ and $100$ nodes, though, a larger sample size, i.e. $n \geq 100,000$, is required to infer d-separations at a reasonable rate. As we will see in Section \ref{section:sage-experiments}, the runtime for graph learning is negligible in the context of $d$-SAGE estimation.

We note that there is no well-defined threshold for the minimal F1 score that would be required for SAGE estimation to benefit from causal structure learning because different error types have distinct consequences. While incorrectly inferred $d$-separations may lead to biased estimates, non-detected $d$-separations only reduce the benefit of CSL in terms of reduced runtime. Importantly, our simulation results in Figure \ref{fig:confusion} show virtually no false discoveries (cases where there is no CI in the underlying distribution but a $d$-separation is inferred) yet some false-negative instances, which leads to fewer skipped evaluations of $\Delta_{j|S}$ than warranted. This result is in accordance with the reasoning presented in Section \ref{subsection-csl}. As such, the use of CSL is a conservative approach to the inference of CIs. Note that for the data used in the benchmark, the ground-truth graph is known and the Markov property and faithfulness hold, such that $d$-separations indeed coincide with statistical independence. 

\subsection{Evaluating Efficiency and Accuracy of \texorpdfstring{$d$}{d}-SAGE\label{section:sage-experiments}}

\begin{figure*}[ht]
\centering
\vspace{.3in}
\includegraphics[width=1\textwidth]{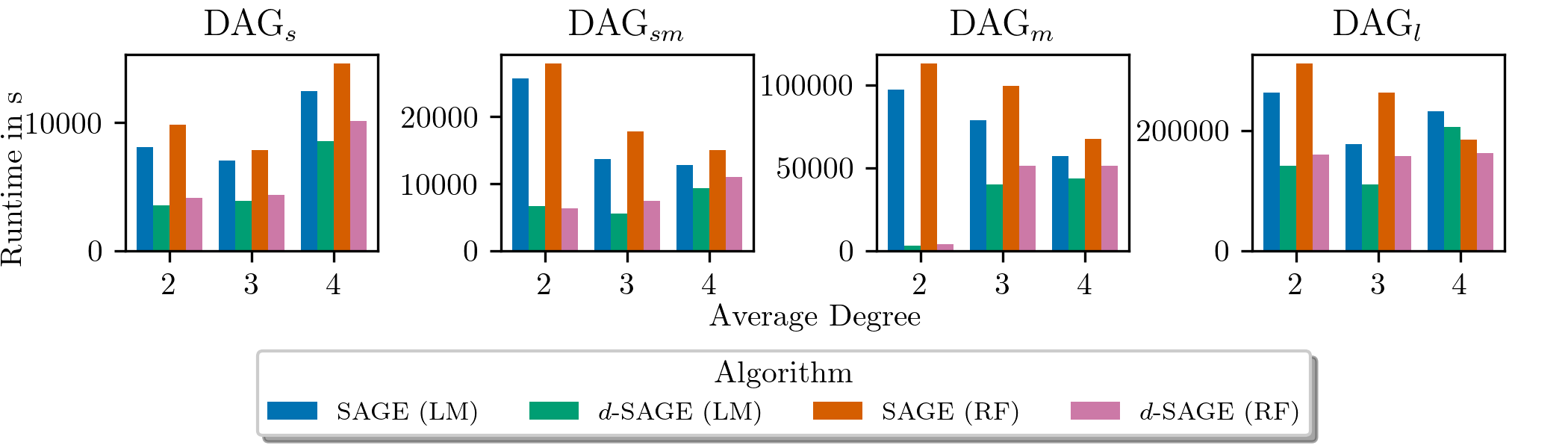}
\caption{Total estimated runtime of one SAGE or $d$-SAGE run for all twelve graphs and linear models (LM) as well as random forests (RF) based on $n=10,000$. \label{fig:sagert}}
\end{figure*}

\begin{figure*}[ht!]
    \centering
        \subcaptionbox{SAGE values and difference between SAGE and $d$-SAGE 
        for the five largest values.}[0.60\textwidth]{
            \centering
                        \includegraphics[width=0.48\linewidth]{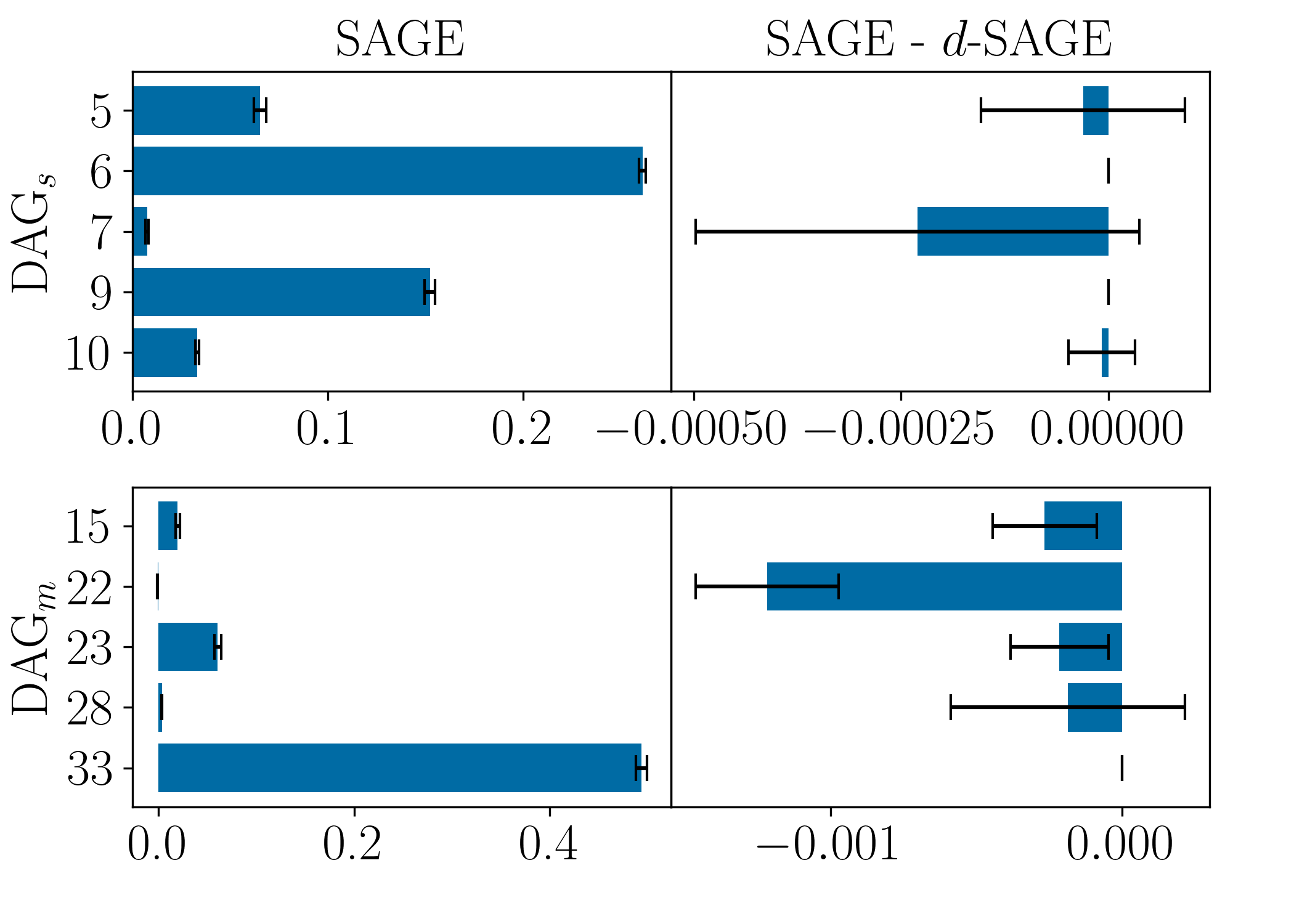}
          \hfill
                        \includegraphics[width=0.48\linewidth]{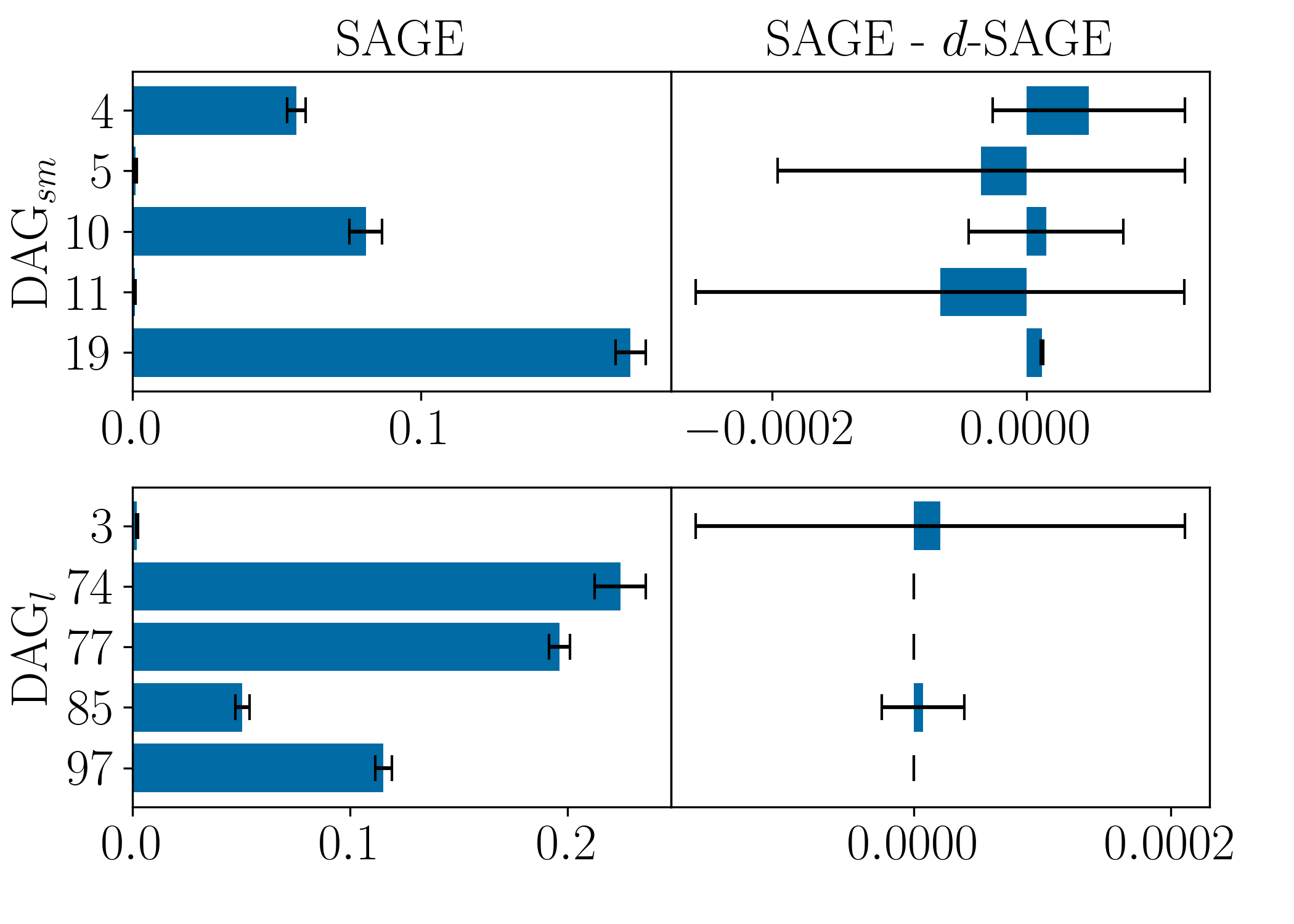}
        }
        \hspace{0.02\linewidth}
        \hfill
        \subcaptionbox{Boxplots showing the distribution of $\Delta_{j|S}$ for the skipped surplus evaluations. \label{fig:box}}[0.36\textwidth]{
            \centering
                \includegraphics[width=0.99\linewidth]{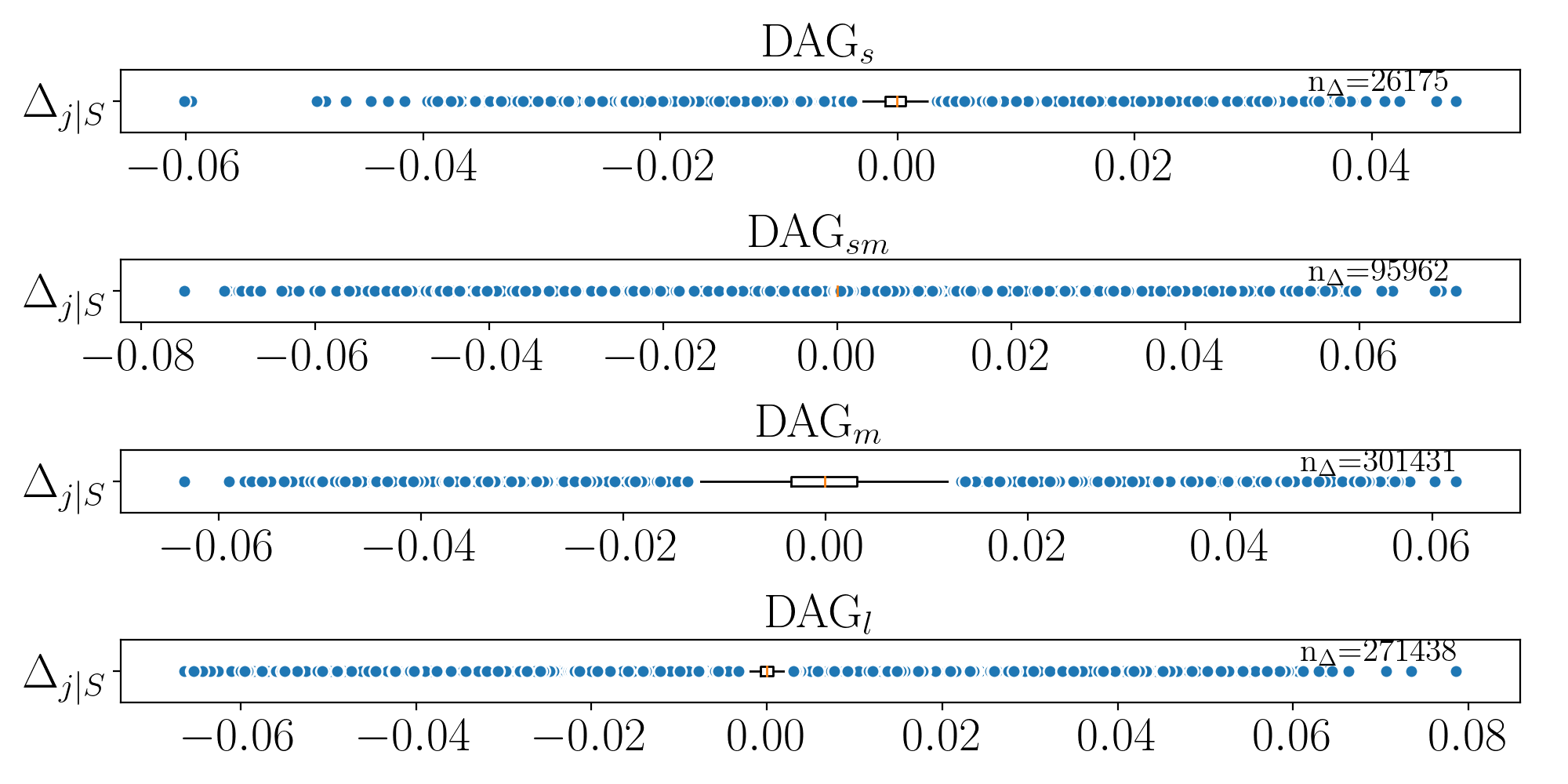}
                \vspace{0.15cm}
        }
        \caption{Results on the approximation quality of $d$-SAGE based on DAGs with average degree $2$ for optimal models (LM). Based on five ($d$-)SAGE estimates. \label{fig:d-sage-accuracy}}
\end{figure*}

In the benchmark study in Section \ref{section:csl-benchmark} we highlight the capability of structure learning to efficiently yet conservatively estimate $d$-separations as equivalents to CIs. We now evaluate $d$-SAGE regarding its efficiency and its accuracy.


\subsubsection{Setup}

To evaluate $d$-SAGE in practice, a linear model (LM) and a random forest (RF) are fitted to each of the twelve datasets (using the \textit{scikit-learn} implementation with default settings \citep{scikit-learn}). 
As loss function, the mean squared error (MSE) is used for either of them. Hence, the linear model (LM) falls into the category of optimal models required for the theoretical justification. The RF model serves as a sanity check for a high-performing, but not optimal model (cf. Appendix \ref{appendix-sage} for the model performances). 
For a fair comparison, we compare $d$-SAGE and SAGE based on the exact same feature orderings. This also allowed us to compare the skipped evaluations of $\Delta_{j|S}$, that are set to zero, to their estimated counterparts that should be very close to zero. Overall, we estimated SAGE and $d$-SAGE values five times for each setup (graph + model). We used the same synthetic datasets for the evaluation as in Section \ref{section:csl-benchmark}.

\subsubsection{Results}
We find that $d$-SAGE indeed speeds up SAGE approximation as expected. More specifically, the estimated runtime\footnote{The complete SAGE estimation was performed on multiple different machines. For a fair evaluation of runtime, we relied on estimates that were performed on the same CPU (Intel Core i7-8700K Desktop CPU): Either approach was conducted using 100 permutations that were the same for SAGE and $d$-SAGE and runtime multiplied by the factor $\frac{n_{\pi}}{100}$, where $n_{\pi}$ is the number of permutations after which one SAGE run converged. For convergence behaviour see Appendix \ref{appendix-convergence}.} decreases by a rate that is approximately equal to the share of CIs w.r.t. the model target (cf. Appendix \ref{appendix-graphs}) for both model classes across all graphs (cf. Figure \ref{fig:sagert}). 
Furthermore, $d$-SAGE manages this speedup without distorting the estimates. Note that we do not include graph learning runtime in Figure \ref{fig:sagert} since it required between 0.06 seconds (DAG$_s$ with average degree 2) and 39.86 seconds (DAG$_l$ with average degree 4) and hence is negligible in this context.

\paragraph{Linear Model} {Figure \ref{fig:d-sage-accuracy} (a) displays the five SAGE values with the largest absolute value for the four graphs with an average degree of two along with the respective difference between the SAGE and $d$-SAGE estimates. Overall, the differences are about three orders of magnitude smaller than the original SAGE values, i.e. typically lie beneath one per cent. Even the most pronounced difference for variable 7 in DAG$_s$ only amounts to approximately $2.7$ per cent of the SAGE value of approximately $0.007$. We find no further striking differences in the remaining SAGE values that identified important features, i.e. those with the largest absolute SAGE values. Features deemed unimportant by SAGE values are detected as such by $d$-SAGE. Noteworthy, some $d$-SAGE estimates are equal to zero if the feature of interest is conditionally independent of the target given all (sampled) coalitions. Here, we argue that we bias observational SAGE values towards zero, which for truly independent features is closer (or equal) to the 'true-to-the-data' estimate that we get with the optimal predictor and infinite data.

Figure \ref{fig:d-sage-accuracy} (b) displays every $\Delta_{j|S}$ value, which was derived from a conditionally independent feature that was detected as such and thus set equal to zero in $d$-SAGE approximation. We see clearly that most values are very close to zero, as mirrored by the narrow boxes, which underlines the usefulness of our approach.}

\paragraph{Random Forest}{
To test the sensitivity of the results, we replicated the exact same study using a high-performing but not optimal RF regressor (instead of the optimal LM). While the runtime savings are the same as for the LM, deviations of $d$-SAGE values from the original estimates are slightly more pronounced (cf. Appendix \ref{appendix-sage}). The results indicate that our approach is also useful for close to optimal models.
}

\subsection{Real-world Application}
\label{subsec:experiments:real-world-application}

To show the usefulness of $d$-SAGE in practice, we applied the approach to drug consumption data from the UCI ML repository \citep{Dua:2019}. The target "Nicotine consumption" was predicted using logistic regression relying on twelve explanatory variables in a dataset with sample size $n=1885$. Graph fitting was conducted with the TABU search algorithm and took $0.035$ seconds. SAGE estimation for five different runs took approximately 12h14min\footnote{All calculations were run on an Intel Core i7-8700K Desktop CPU}. To derive $d$-SAGE values, we did not rerun the estimation relying on $d$-SAGE but simply replaced the respective $\Delta_{j|S}$ that pertained to a $d$-separation in the fitted graph in the output (that included all such $\Delta_{j|S}$) with zero. We found approximately $38$ per cent such $\Delta_{j|S}$ values that can be skipped which warrants an (almost) equally large relative speedup.

\begin{figure}[ht]
    \centering
        \includegraphics[width=0.99\linewidth]{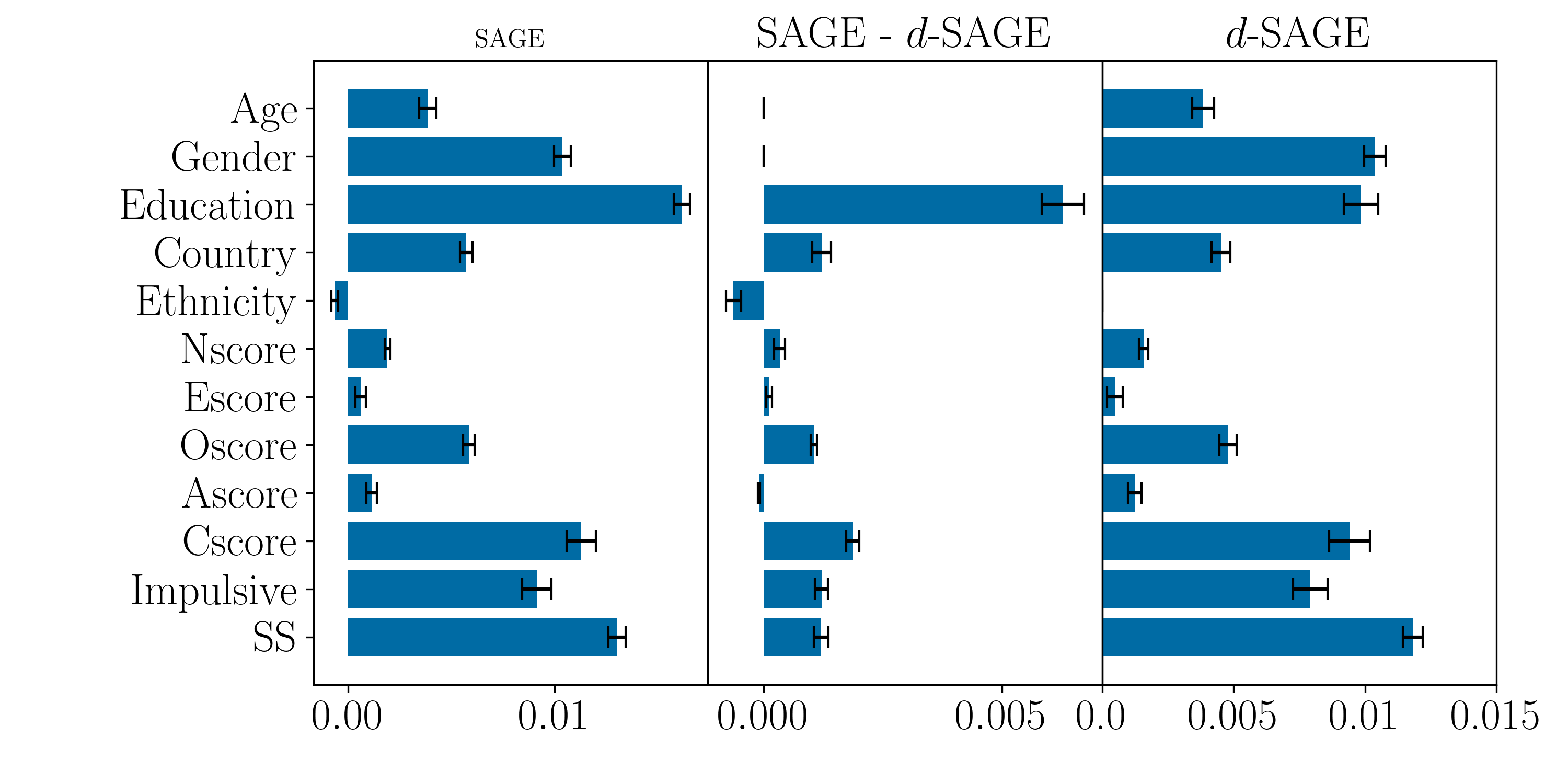}
    \caption{SAGE values, difference between SAGE and $d$-SAGE and $d$-SAGE values for drug consumption data. Based on five ($d$-)SAGE estimates.}
        \label{fig:sage-app-drug}
\end{figure}

\begin{figure}[ht]
    \centering
        \includegraphics[width=0.9\linewidth]{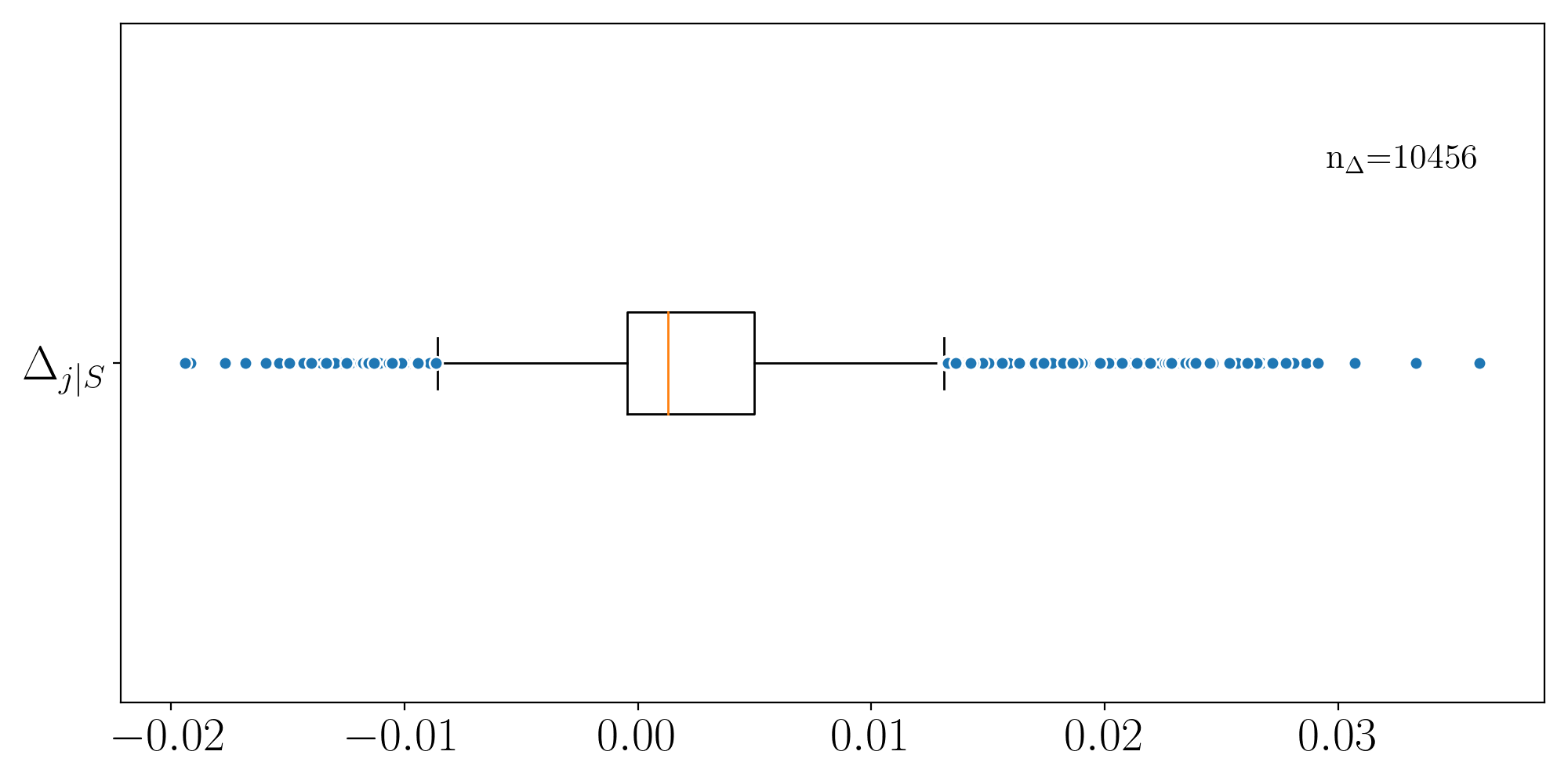}
    \caption{Boxplots showing the distribution of $\Delta_{j|S}$ for the skipped surplus evaluations. Based on five ($d$-)SAGE estimates.}
        \label{fig:delta-app-drug}
\end{figure}

Figure \ref{fig:sage-app-drug} shows that $d$-SAGE values are mostly in accordance with the original SAGE estimate. From the important variables, only \textit{'Education'} has a markedly distinct $d$-SAGE value as it is reduced by about a third compared to the SAGE estimate. Yet, it is still assigned relatively high importance. The efficacy of $d$-SAGE in practice is further highlighted by the $\Delta_{j|S}$ values that hover around zero, as shown in Figure \ref{fig:delta-app-drug}.    

\section{DISCUSSION}

\paragraph{Model Optimality}{
Conditional SAGE values are particularly appealing for scientific inference, i.e. to learn about the data \citep{chen2020true,covert2020understanding}. Therefore, in general, accurate predictors are required \citep{molnar2022general}. However, the requirement is of increased importance for $d$-SAGE since if the assumption of model optimality is violated, the interpretation may be further biased by skipping the evaluation of non-zero surplus contributions (Theorem \ref{theorem:surplus-zero}).
}


\paragraph{Assumptions for CSL}{
CSL is enabled by causal sufficiency, the Markov property and faithfulness \citep{Peters2017}. The assumptions ensure that all relevant variables are observed, and that CIs in the data coincide with $d$-separations in the true causal graph (which we assume to be a DAG). We conjecture that violations of these assumptions are not vital for our approach since learning the true causal graph is not the goal. Instead, we are only interested in learning the graph to encode (conditional) independencies present in the observational distribution (irrespective of which causal mechanism they stem from). DAGs learned by HC being a minimal I-Map of the underlying distribution makes it suitable for probabilistic inference of CIs without guarantees of a correct graph or the number of CIs uncovered.

Nevertheless, practitioners should carefully assess the assumptions before applying $d$-SAGE. In the presence of latent confounders or cyclic assignment, for example, one may consider other concepts, such as $m$-separation and $\sigma$-separation (cf. \cite{bongers2021causal}). Moreover, it is advisable to perform sanity checks on whether skipped surplus contributions are actually evaluated to zero.
}





\paragraph{Use of Score-based CSL}{The analysis was restricted to the use of score-based CSL because of its efficiency. HC is particularly appealing since it infers a minimal I-Map of the underlying distribution as explained in Section \ref{subsection-csl}, and TABU performed well empirically. However, inference of CIs is not limited to those techniques. Graph learning can be performed with an algorithm of choice and under consideration of the assumptions employed, as explained above. Moreover, the rationale behind our approach is to replace CI testing by CSL. Partial correlation tests, for example, are considerably less efficient than $d$-separation queries (cf. Appendix \ref{appendix-parr_corr}) and thus would require a larger number of CIs to achieve a speedup of SAGE.
}

\section{CONCLUSION}

We proposed $d$-SAGE, a method that exploits the dependence structure in the data to speed up SAGE estimation.
More specifically, we observe that conditional independence in the data implies that the corresponding surplus contribution can be directly evaluated to zero.
We modify the ordering based SAGE approximation algorithm to first learn the dependence structure in the data using CSL algorithms and to then skip surplus contribution evaluations if the graph encodes a CI.
Errors in the learned graph may either slow down convergence (if CIs are not discovered) or bias the result towards zero (in case of false discoveries).
However, in our experiments, there were nearly no false discoveries, such that the resulting estimates for features that were not conditionally independent given every coalition essentially converged to the same values as the original SAGE approximation algorithm.
Furthermore, the CSL algorithms were able to uncover most CIs, such that we observe significant performance gains. 
As such, given a fixed computational budget, the efficiency gains of $d$-SAGE can enable a more accurate estimation of SAGE values than the approximation algorithm proposed by \cite{covert2020understanding}.
In future work, it would be interesting to combine $d$-SAGE with the permutation sampling by \cite{mitchell2022sampling} and to assess whether the results can be translated to other Shapley based interpretability methods such as SHAP.

\subsubsection*{Acknowledgements}
This project is supported by the German Federal Ministry of Education and Research (BMBF). The computational results presented have been achieved in part using the Vienna Scientific Cluster (VSC). We thank Bernd Bischl for his advice on benchmark design.

\bibliography{main.bib}
\appendix
\onecolumn


\section{PROOF OF THEOREM \ref{theorem:surplus-zero}}
\label{sec:theorem:surplus-zero}

\begin{reptheorem}{theorem:surplus-zero}

For $\ell$ being cross-entropy loss or the mean-squared error, $f^*$ the respective optimal predictor and $\nu_{\ell,f^*}$ the corresponding SAGE value function, it holds that
\begin{align*}
    X_j \perp Y | \textbf{X}_S \Rightarrow \nu_{\ell,f^*}(\textbf{X}_{S \cup j}) - \nu_{\ell,f^*}(\textbf{X}_S) = 0.
\end{align*}

\end{reptheorem}

\begin{proof}

%
%
%
%
\textit{Mean Squared Error:} \cite{covert2020understanding} show that for $\ell$ being the mean squared error and $f^*$ the corresponding optimal predictor it holds that:
$$\nu(\textbf{X}_{S \cup j}) - \nu(\textbf{X}_{S}) = \mathbb{E}[\text{Var}(Y|\textbf{X}_{S})] - \mathbb{E}[\text{Var}(Y|\textbf{X}_{S \cup j})]$$
Under conditional independence $Y \perp X_j | \textbf{X}_{S}$ it follows that
\begin{align*}
\mathbb{E}[\text{Var}(Y|\textbf{X}_{S \cup j})] &= \mathbb{E}[\mathbb{E}[\text{Var}(Y|\textbf{X}_{S \cup j})|\textbf{X}_{S}]]\\
&= \mathbb{E}[\text{Var}(Y|\textbf{X}_{S})]
\end{align*}
and consequently $Y \perp X_j | \textbf{X}_{S} \Rightarrow \nu(\textbf{X}_{S \cup j}) - \nu(\textbf{X}_{S})  = 0$.\\

\textit{Cross Entropy:} \cite{covert2020understanding} show that given cross entropy as loss and the corresponding loss optimal predictor $f^*$ it holds that:
$$\nu(\textbf{X}_{S \cup j}) - \nu(\textbf{X}_{S}) = I(Y;X_j|\textbf{X}_{S})$$
Mutual information $I(Y;X_j|\textbf{X}_{S})$ is zero if and only if $Y \perp X_j | \textbf{X}_{S}$. Consequently $\nu(\textbf{X}_{S \cup j}) - \nu(\textbf{X}_{S}) = 0 \Leftrightarrow X_j \perp Y | \textbf{X}_{S}$.\\
\end{proof}

\section{SAGE VALUE PROPERTIES}

As mentioned in Section \ref{section-background}, SAGE values satisfy certain fairness properties that are deduced from those valid for Shapley values \citep{covert2020understanding}. While not explicitly named after the Shapley value properties (\textit{efficiency}, the \textit{dummy property}, \textit{symmetry}, \textit{monotonicity}, \textit{linearity}) we employ these terms for the SAGE properties for simplicity:

\begin{enumerate}

    \item \textit{Efficiency}: $\sum_{j=1}^d \phi_j(\nu) = \nu(\textbf{X})$, where \textbf{X} is the set of all features.
    
    \item \textit{Dummy property}: $\phi_j(\nu) = 0 $ if $X_j \perp \hat{f}(\textbf{X}) | \textbf{X}_S$ for all $S \subseteq \{1,...,d\} \setminus j$.
        
    \item \textit{Symmetry}: $\nu(\textbf{X}_{S \cup j}) = \nu(\textbf{X}_{S \cup i})$ for two variables $X_j$ and $X_i$ with a deterministic relationship.
    
    \item  \textit{Monotonicity}: For two target variables $Y$, $Y'$ and corresponding models $\hat{f}$, $\hat{f'}$: $\phi_j(\nu_{\hat{f}}) \geq \phi_j(\nu_{\hat{f'}})$ if $\nu_{\hat{f}}(\textbf{X}_{S \cup j}) - \nu_{\hat{f}}(\textbf{X}_{S}) \geq \nu_{\hat{f'}}(\textbf{X}_{S \cup j}) - \nu_{\hat{f'}}(\textbf{X}_{S})$ for all $S \subseteq \{1,...,d\} \setminus j$.
    
    \item From \textit{Linearity}: $\phi_j(\nu) = \mathbb{E}_{\textbf{X}, Y}[\phi_j(\nu_{\hat{f},x,y}) ]$, where $\phi_j(\nu_{\hat{f},x,y})$ is the Shapley value of the game $\nu_{\hat{f},x,y}(\textbf{X}_S) = \ell(\hat{f}_{\emptyset}(\textbf{X}_{\emptyset}),y) - \ell(\hat{f}_{S}(\textbf{X}_S),y) $
    
    \item SAGE values are invariant to invertible mappings applied to the input, e.g. they are the same for original input data and and their log values.
    
\end{enumerate}

\clearpage
\section{GRAPH BENCHMARK} \label{appendix-graphs}

In this section, we provide detailed information about the graphs employed in Section \ref{section-experiments}, the graph learning algorithms and the graph benchmark. Additionally, we present results derived from the HC algorithm for CSL.

\subsection{Overview of Graphs}

In Table \ref{table-graphs_overview} we provide an overview of all twelve graphs used in Section \ref{section-experiments}, the randomly sampled target, the adjacency degree of the target and the share of $d$-separations w.r.t. the target. This gives further insight into the relation of graph sparsity, degree of target and share of $d$-separations. The latter can be regarded as the potential relative runtime decrease for SAGE approximation.

\begin{table}[ht]
\caption{Overview of all twelve graphs used in Section \ref{section-experiments}, the randomly sampled target, the adjacency degree of the target and the share of $d$-separations w.r.t. the target.} \label{table-graphs_overview}
\begin{center}
\begin{tabular}{l c c c}
\textbf{GRAPH (AVG. DEGREE)}  & \textbf{TARGET} & \textbf{DEGREE OF TARGET} & \textbf{SHARE OF} $\perp_{\mathcal{G}}$ \\
\hline \\
DAG$_s$(2)      & 8 & 2 & 0.556  \\
DAG$_s$(3)      & 1 & 2 & 0.357  \\
DAG$_s$(4)      & 1 & 4 & 0.283  \\
DAG$_{sm}$(2)   & 17 & 1 & 0.765  \\
DAG$_{sm}$(3)   & 2 & 1 & 0.623  \\
DAG$_{sm}$(4)   & 16 & 4 & 0.185  \\
DAG$_m$(2)      & 4 & 1 & 0.961  \\
DAG$_m$(3)      & 32 & 5 & 0.556 \\
DAG$_m$(4)      & 2 & 3 & 0.274  \\
DAG$_l$(2)      & 4 & 3 & 0.632  \\
DAG$_l$(3)      & 66 & 3 & 0.552  \\
DAG$_l$(4)      & 66 & 7 & 0.151  \\

\end{tabular}
\end{center}
\end{table}

Table \ref{table-csl} shows the hyperparameter settings used for CSL relying on the \textit{bnlearn} package \citep{scutari2010bnlearn} for R \citep{rcore2022}.

\begin{table}[ht]
\caption{Hyperparameters Used for Graph Learning} \label{table-csl}
\begin{center}
\begin{tabular}{l c}
\textbf{ALGORITHM} & \textbf{HYPERPARAMETERS} \\
\hline \\
     HC &  Max. iterations $\infty$, max. in-degree: $\infty$; score: BIC  \\
        
     TABU &  Size of list: 10; Max. iterations $\infty$, max. in-degree: $\infty$; score: BIC  \\
\end{tabular}
\end{center}
\end{table}

\subsection{MC Sampling for \texorpdfstring{$d$}{d}-separation Inference}

In Algorithm \ref{alg:mcdsep} we explicate how we inferred the number of true positive, false positive, true negative and false negative $d$-separations within an estimated graph and especially how we dealt with the exponential number of potential conditioning sets for the larger graphs.

\normalem 
\begin{algorithm}[ht] 
\small
\SetAlgoLined

\textbf{Input:} True graph $\mathcal{G}^*$ and estimated graph $\mathcal{G}$ over node set $\{X_1, X_2,... X_d, Y \}$ with target node $Y$; Number of MC samples $n_{mc}$\\ 
\textbf{Output:} True positives, true negatives, false positives and false negatives for inferred $d$-separations in $\mathcal{G}$: TP, TN, FP, FN  \\

   Set $TP=TN=FP=FN=0$ \\
   \For{$m=1,...,n_{mc}$}{
       Randomly draw a node $X_j$ from $\{X_1, X_2,... X_d \}$ \\
       Randomly draw size $n_s$ of conditioning set $\textbf{X}_S$ from discrete probability distribution $P(n_s = i) = \frac{{d-1\choose i}}{2^{d-1}}$, $i \in \{0,...,d-1\}$ \\
       Randomly draw elements $X_i$, $i=1,...n_s$, from  $\{X_1, X_2,... X_d \} \setminus X_j$ without replacement and set $\textbf{X}_S = \{X_i \}_{i=1,...,n_s}$ \\
       \eIf{$X_{j} \perp_{\mathcal{G}^*} Y | \textbf{X}_S$}{\eIf{$X_{j} \perp_{\mathcal{G}} Y | \textbf{X}_S$}{TP = TP+1}{FN = FN+1}}{\eIf{$X_{j} \not\perp_{\mathcal{G}} Y | \textbf{X}_S$}{TN = TN+1}{FP = FP+1}}
    }
\textbf{Return:} TP, TN, FP, FN      
\caption{Monte Carlo Sampling for \texorpdfstring{$d$}{d}-separation Inference}\label{alg:mcdsep}
\end{algorithm}

\ULforem 

\subsection{Results - HC}

\begin{figure*}[ht]
\centering
\begin{subfigure}[t]{0.44\textwidth}
  \centering
  \includegraphics[width=0.88\linewidth]{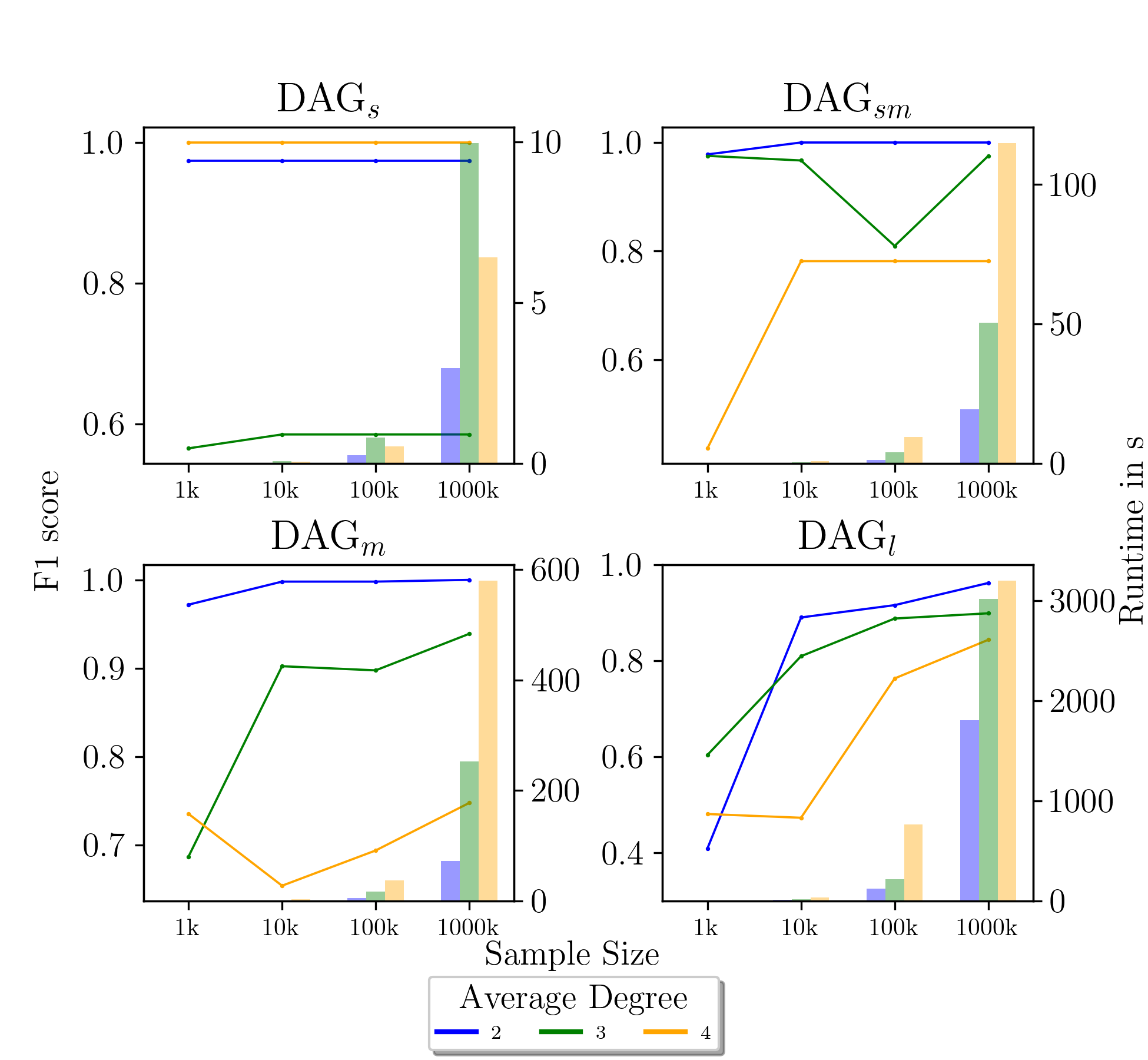}
  \caption{F1 scores for $d$-separation (lines, left y-axes) and runtime of graph learning (bars, right y-axes) using HC depending on sample size. \label{fig:cslruntimehc}}
\end{subfigure}%
\hfill
\begin{subfigure}[t]{0.54\textwidth}
  \centering
  \includegraphics[width=0.99\linewidth]{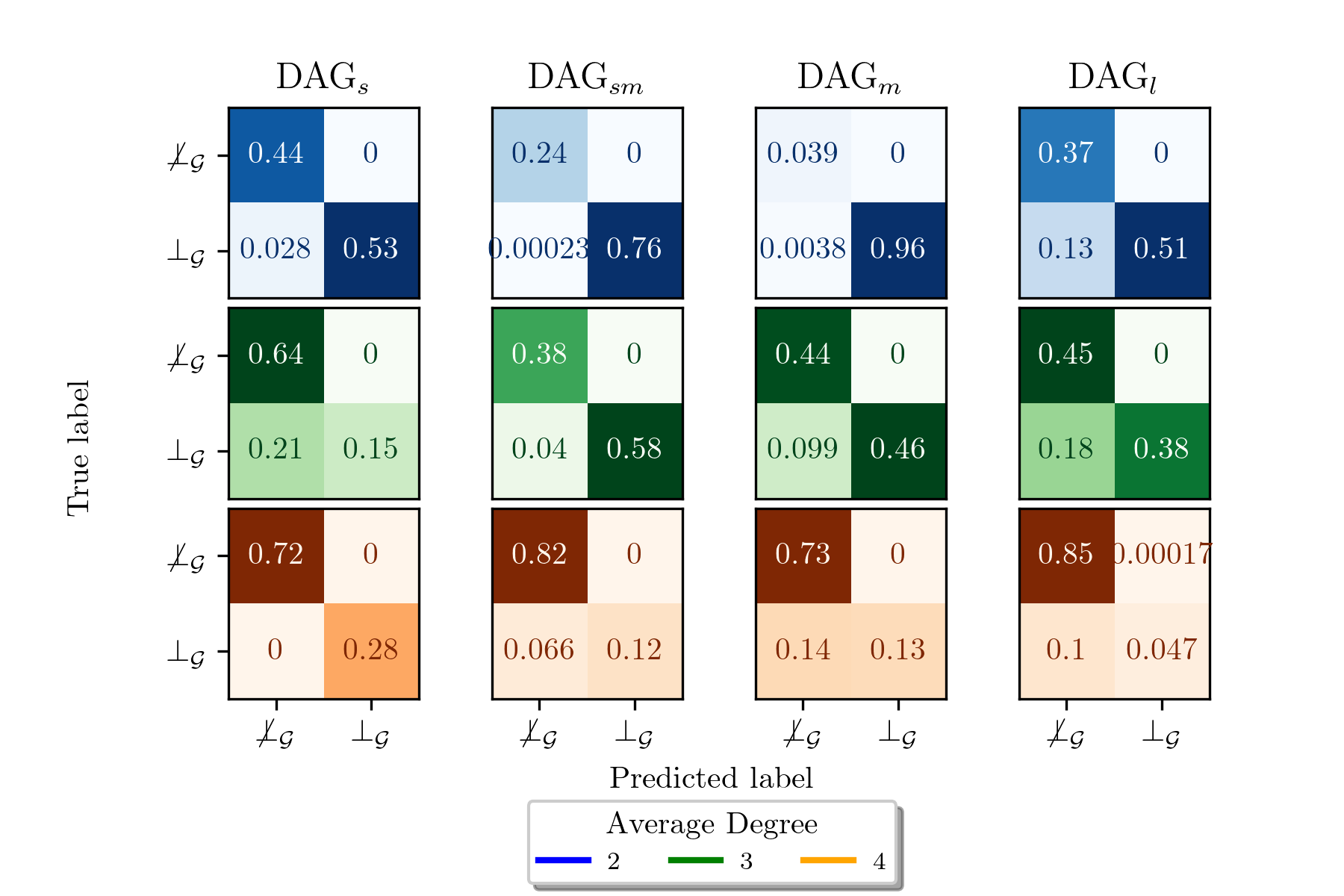}
 \caption{Confusion matrix for true and predicted $d$-connections ($\not \perp_{\mathcal{G}}$) and $d$-separations ($\perp_{\mathcal{G}}$) based on HC with $n=10,000$  for all twelve graphs. \label{fig:confusionhc}}
\end{subfigure}
\caption{Results from graph learning benchmark for HC algorithm. \label{fig:completehc}}
\end{figure*}

In Figure \ref{fig:completehc} we show the results of the graph learning benchmark for HC in contrast to those from Section \ref{section-experiments}. As HC never performed better but for some experiments worse than TABU, we chose the latter for the use in $d$-SAGE.

\clearpage

\section{SAGE - EXPERIMENTS} \label{appendix-sage}

In this section, we briefly explain the experiment setup and afterwards present missing results. For our analysis, we fitted two models, LM and RF, for every dataset relying on the same targets that were sampled randomly for the analysis of $d$-separations in a graph. We relied on $n_{train}=8000$ for model fitting and $n_{test}=2000$ for model evaluation (the same $n=10000$ data points as used for graph fitting and SAGE inference). We then used the data to estimate SAGE and $d$-SAGE five times, i.e. we were provided five approximations of ($d$-)SAGE for every graph and model, which were then used to provide error bounds. The $\Delta_{j|S}$ plots rely on skipped evaluations of each of these runs.

In Table \ref{tab:contmodel} we provide performance measures of the models and in Appendix \ref{app-results} the plots pertaining to experiments not shown in Section \ref{section-experiments} are displayed. Note that Table \ref{tab:contmodel} highlights that RF performs slightly worse than the optimal LM throughout all settings and with regard to the MSE and $R^2$.

\begin{table}[ht]
\caption{Details of Linear Models (LMs) and Random Forests (RF); Random Forests based on 100 Tree Estimators.} \label{tab:contmodel}
\begin{center}
\begin{tabular}{l c c c c c}
\textbf{DATA (AVERAGE DEGREE)} & \textbf{n}$_{train}$; \textbf{n}$_{test}$ & \textbf{MSE}$_{LM}$ & \textbf{R}$^2_{LM}$ & \textbf{MSE}$_{RF}$ & \textbf{R}$^2_{RF}$ \\
\hline \\
                DAG$_{s}$(2) & 8000; 2000   & 0.541 & 0.495 & 0.572 & 0.466\\
        
                DAG$_{sm}$(2) & 8000; 2000  & 0.035 & 0.963 & 0.038 & 0.960 \\
        
                DAG$_{m}$(2) & 8000; 2000   & 0.474 & 0.522 & 0.498 & 0.498 \\
        
                DAG$_{l}$(2) & 8000; 2000   & 0.070 & 0.930 & 0.103 & 0.897 \\

                DAG$_{s}$(3) & 8000; 2000   & 0.382 & 0.616 & 0.480 & 0.517\\
        
                DAG$_{sm}$(3) & 8000; 2000  & 0.072 & 0.926 & 0.078 & 0.921 \\
        
                DAG$_{m}$(3) & 8000; 2000   & 0.089 & 0.914 & 0.174 & 0.832 \\
        
                DAG$_{l}$(3) & 8000; 2000   & 0.065 & 0.938 & 0.082 & 0.922 \\

                DAG$_{s}$(4) & 8000; 2000   & 0.101 & 0.902 & 0.161 & 0.843\\
        
                DAG$_{sm}$(4) & 8000; 2000  & 0.075 & 0.925 & 0.086 & 0.914 \\
        
                DAG$_{m}$(4) & 8000; 2000   & 0.163 & 0.840 & 0.194 & 0.810 \\
        
                DAG$_{l}$(4) & 8000; 2000   & 0.004 & 0.996 & 0.059 & 0.943 \\

\end{tabular}
\end{center}
\end{table}

\subsection{Results - SAGE and \texorpdfstring{$d$}{d}-SAGE \label{app-results}}

In this section we provide the same results as in Section \ref{section-experiments} for all missing setups and both models, LM and RF as well as the top fifteen values for the setup presented in Section \ref{section-experiments}. Overall, we can confirm our findings in the different settings.


\begin{figure*}[ht]
    \centering
        \begin{subfigure}{0.5\textwidth}
            \centering
                \includegraphics[width=0.99\linewidth]{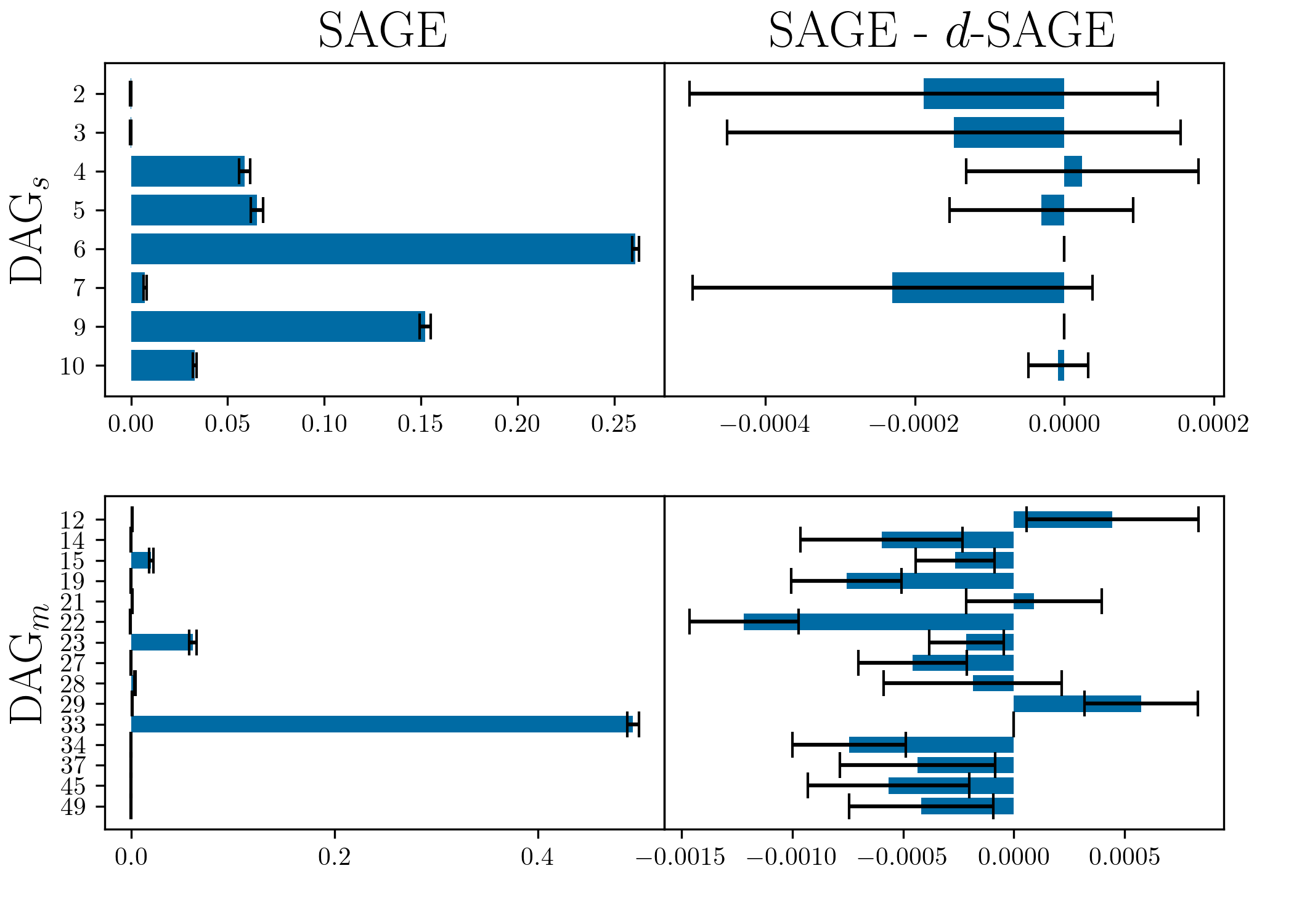}
            \end{subfigure}%
            \hfill
                \begin{subfigure}{0.5\textwidth}
                    \centering
                        \includegraphics[width=0.99\linewidth]{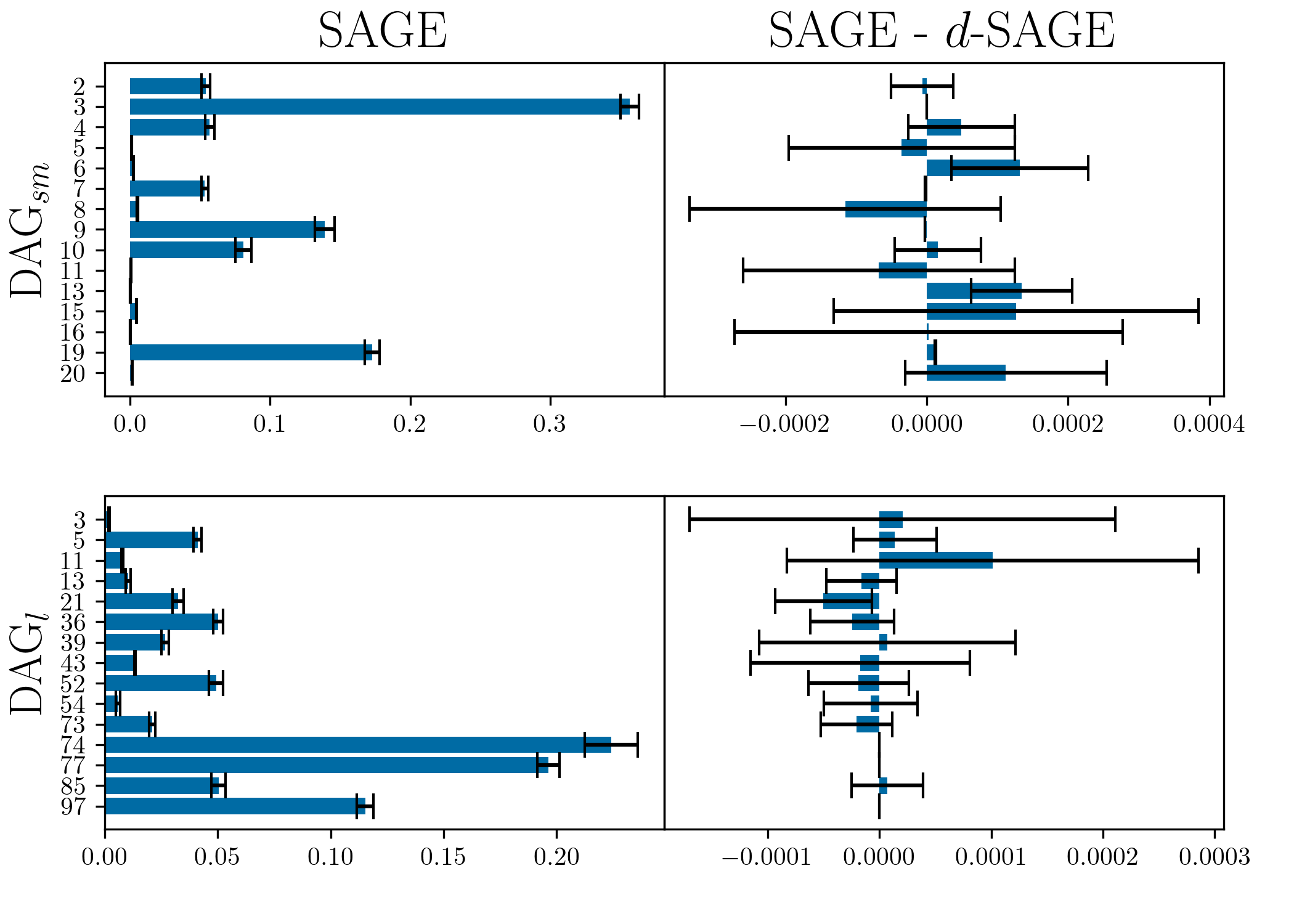}
                \end{subfigure}%
            \caption{SAGE values and difference between SAGE and $d$-SAGE for the fifteen (all for DAG$_s$) largest values for optimal models for DAGs with average degree two. Based on five ($d$-)SAGE estimates.} 
    \end{figure*}

\begin{figure*}[ht]
    \centering
        \begin{subfigure}{0.5\textwidth}
            \centering
                \begin{subfigure}{0.5\textwidth}
                    \centering
                        \includegraphics[width=0.99\linewidth]{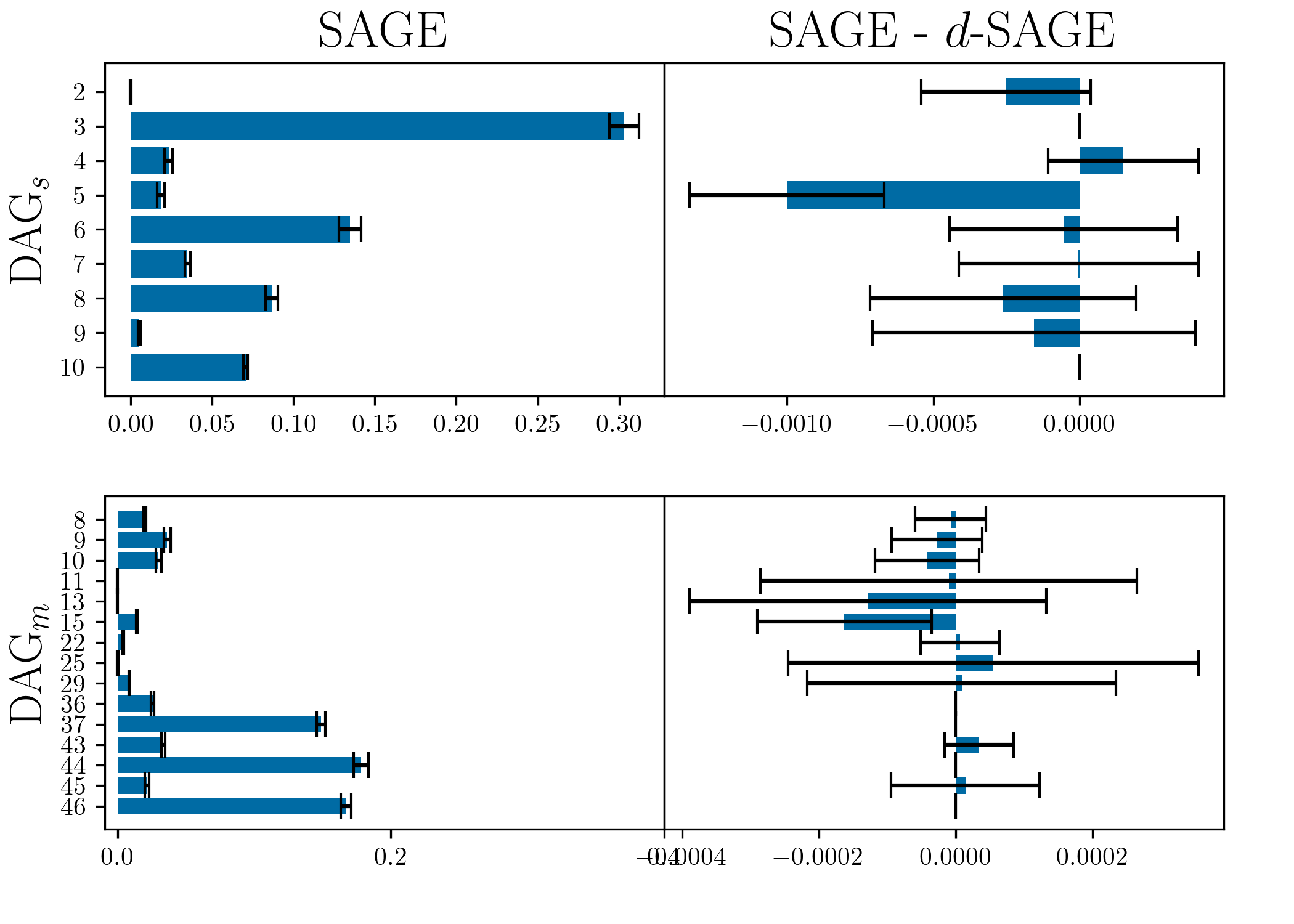}
                \end{subfigure}%
            \hfill
                \begin{subfigure}{0.5\textwidth}
                    \centering
                        \includegraphics[width=0.99\linewidth]{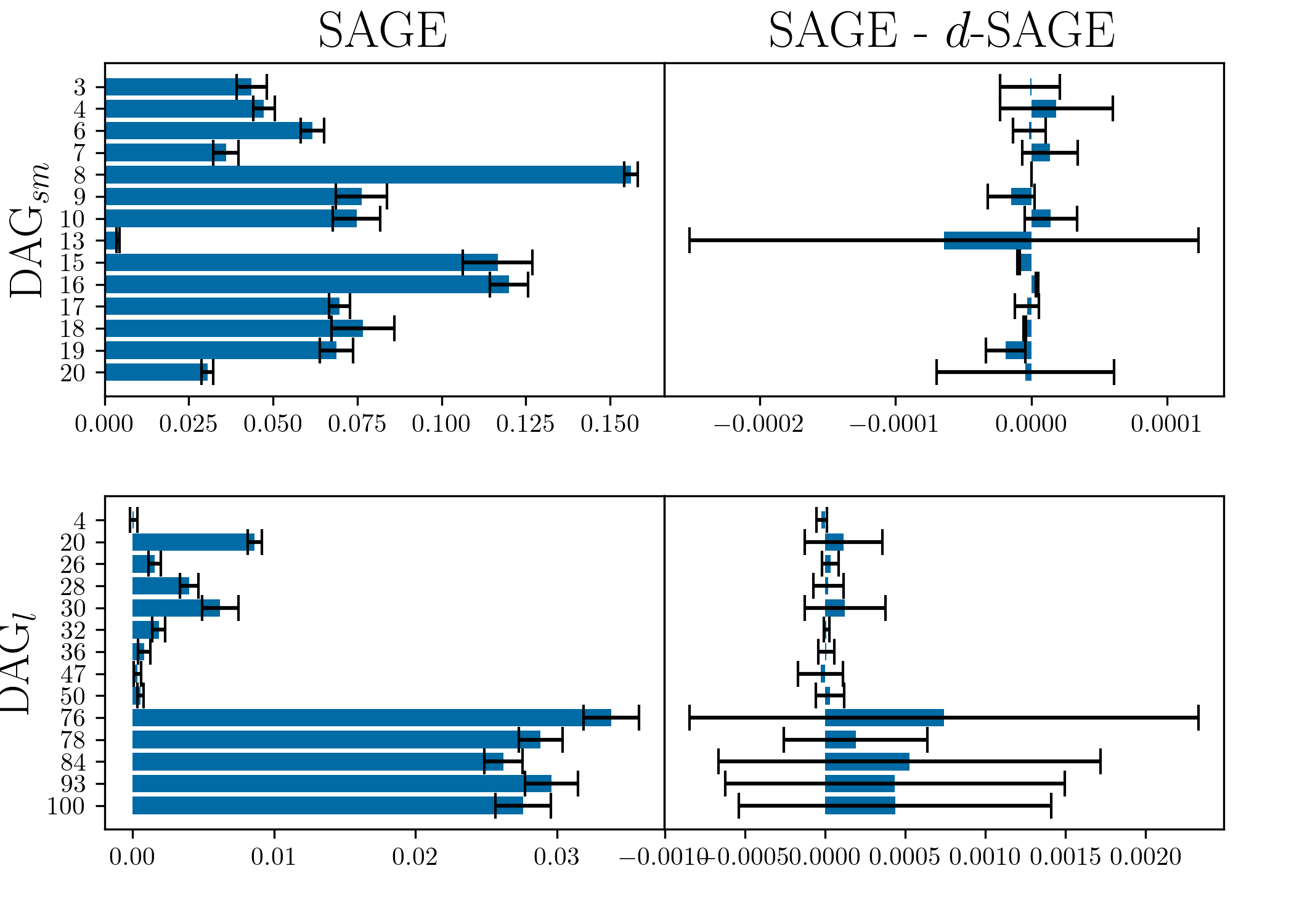}
                \end{subfigure}%
            \caption{SAGE values and difference between SAGE and $d$-SAGE for the fifteen (all for DAG$_s$) largest values for optimal models.}
        \end{subfigure}%
    \hfill
        \begin{subfigure}{0.5\textwidth}
            \centering
                \includegraphics[width=0.99\linewidth]{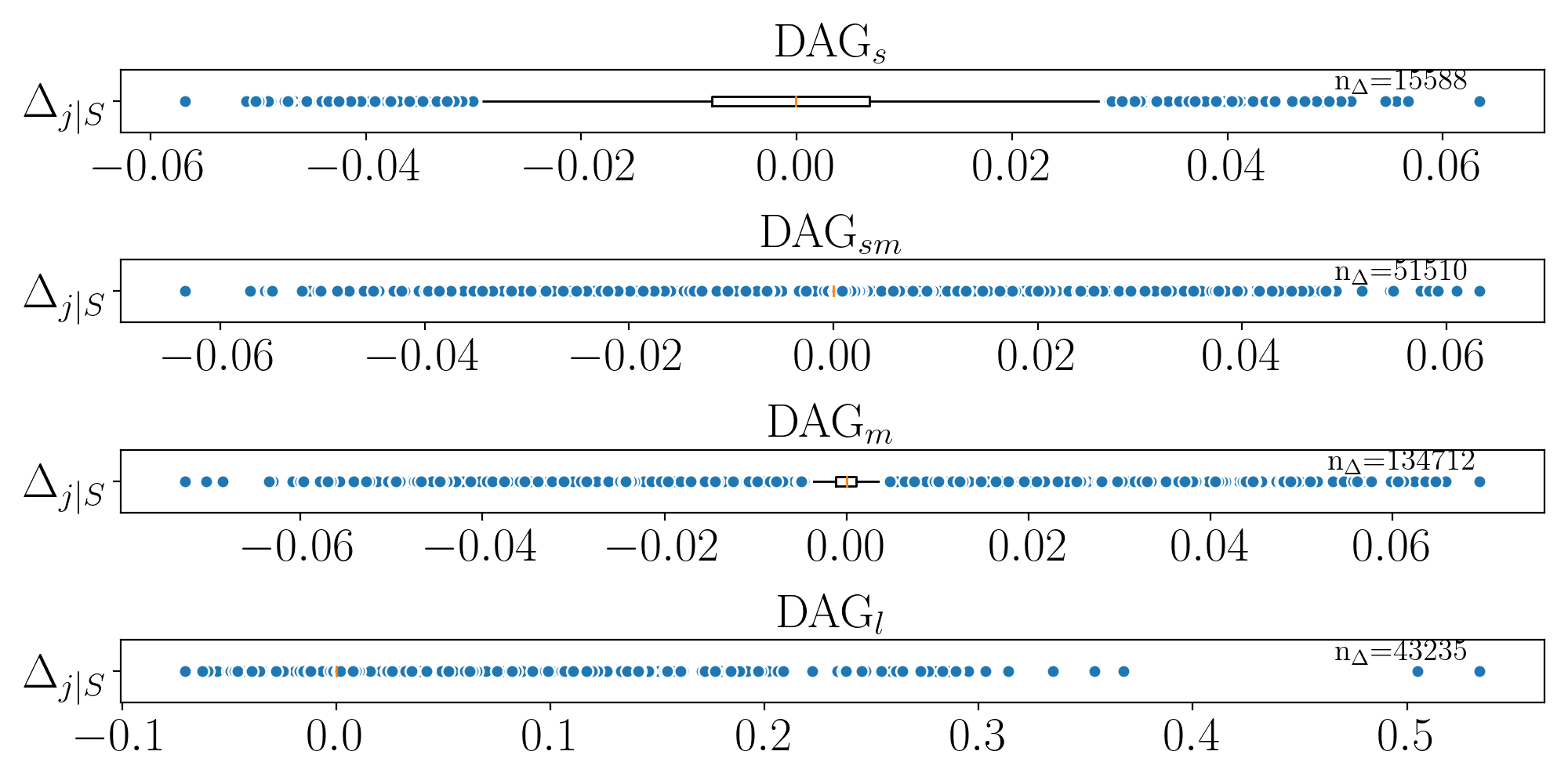}
            \caption{Boxplots showing the distribution of $\Delta_{j|S}$ for the skipped surplus evaluations.}     
        \end{subfigure}%
    \caption{Results on the estimation quality for $d$-SAGE based on each DAG with average degree three and the LM.}
        \label{fig:sage-app-lm3}  
\end{figure*}



\begin{figure*}[ht]
    \centering
        \begin{subfigure}{0.5\textwidth}
            \centering
                \begin{subfigure}{0.5\textwidth}
                    \centering
                        \includegraphics[width=0.99\linewidth]{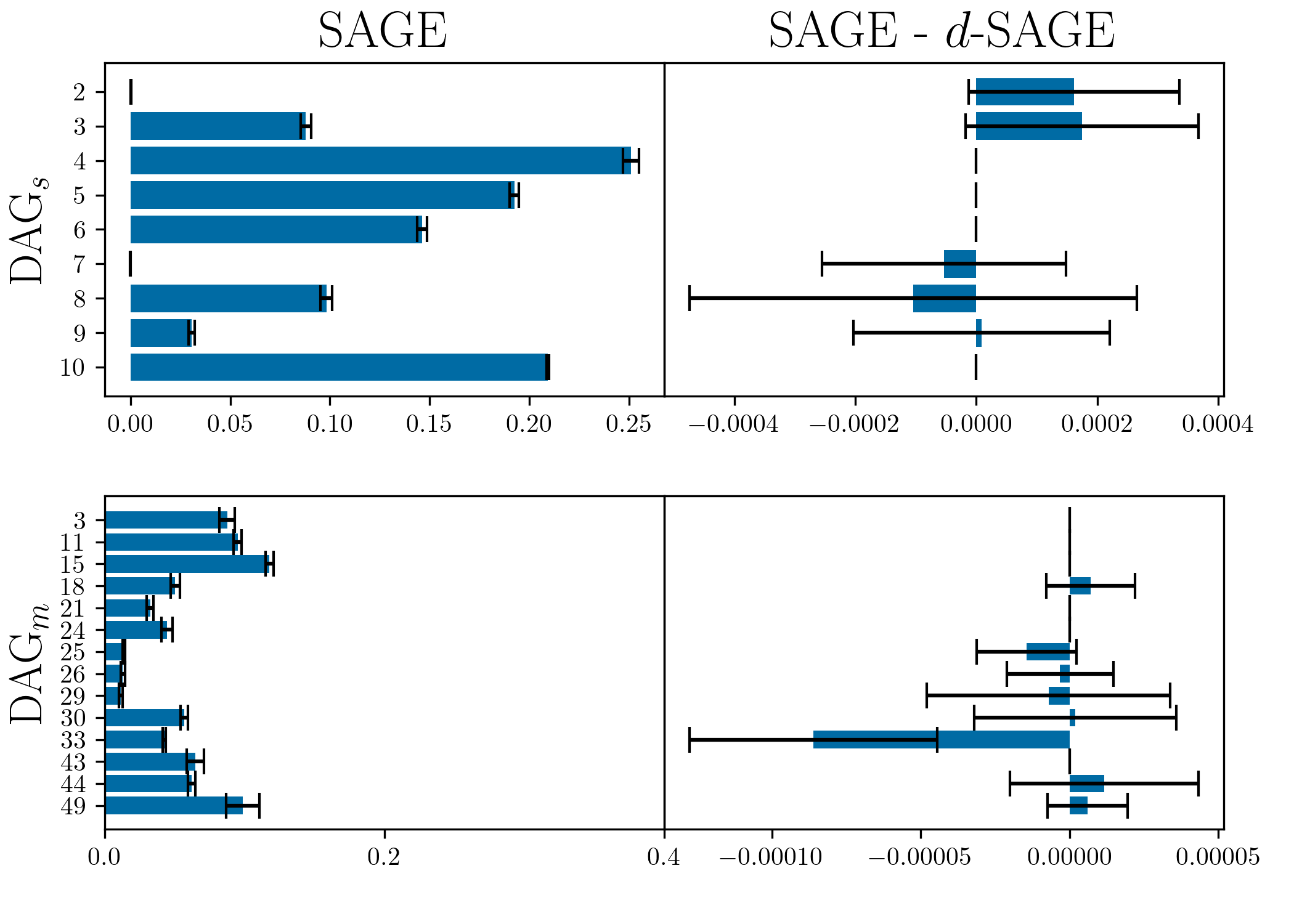}
                \end{subfigure}%
            \hfill
                \begin{subfigure}{0.5\textwidth}
                    \centering
                        \includegraphics[width=0.99\linewidth]{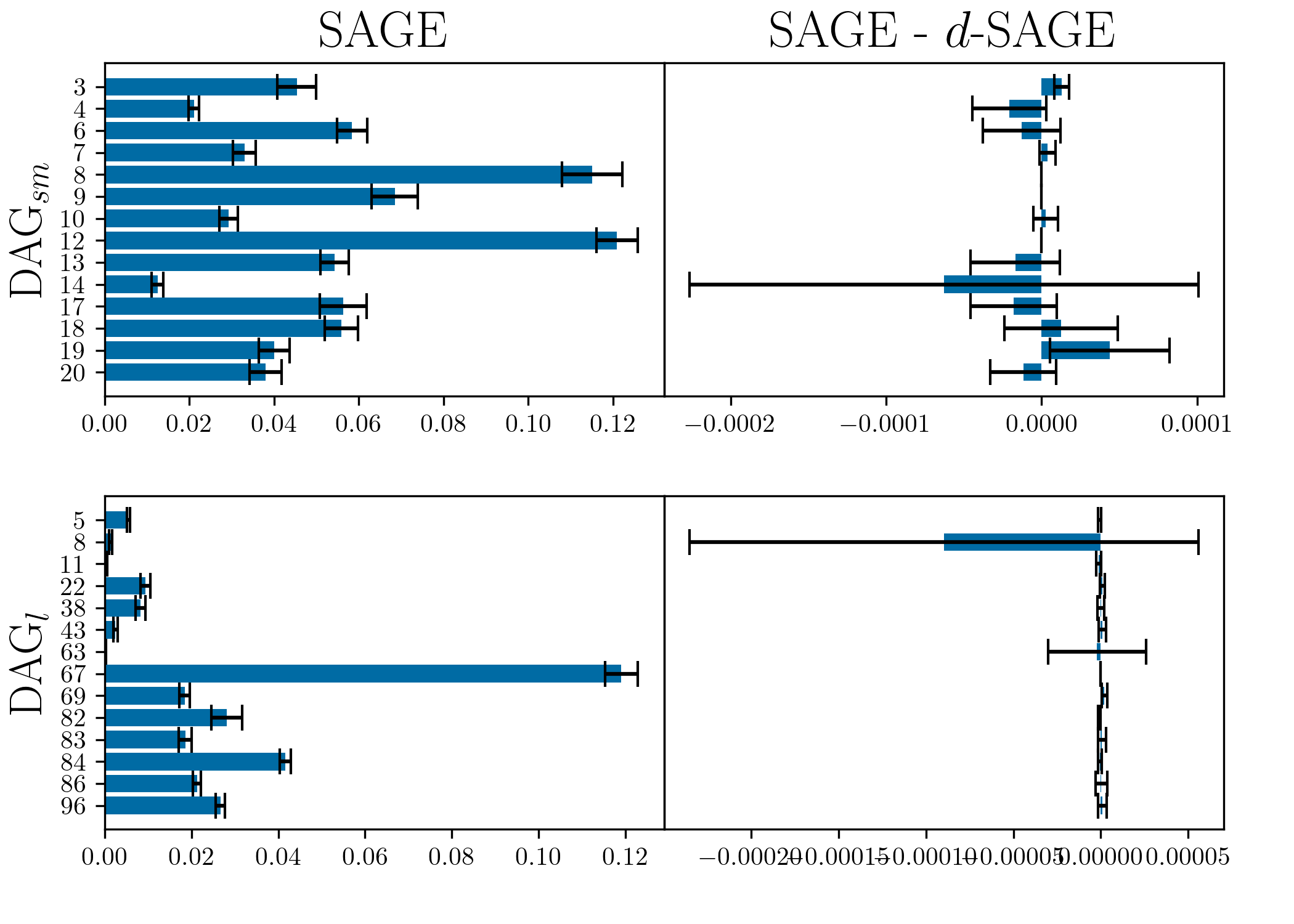}
                \end{subfigure}%
            \caption{SAGE values and difference between SAGE and $d$-SAGE for the fifteen largest (all for DAG$_s$) values for optimal models}
        \end{subfigure}%
    \hfill
        \begin{subfigure}{0.5\textwidth}
            \centering
                \includegraphics[width=0.99\linewidth]{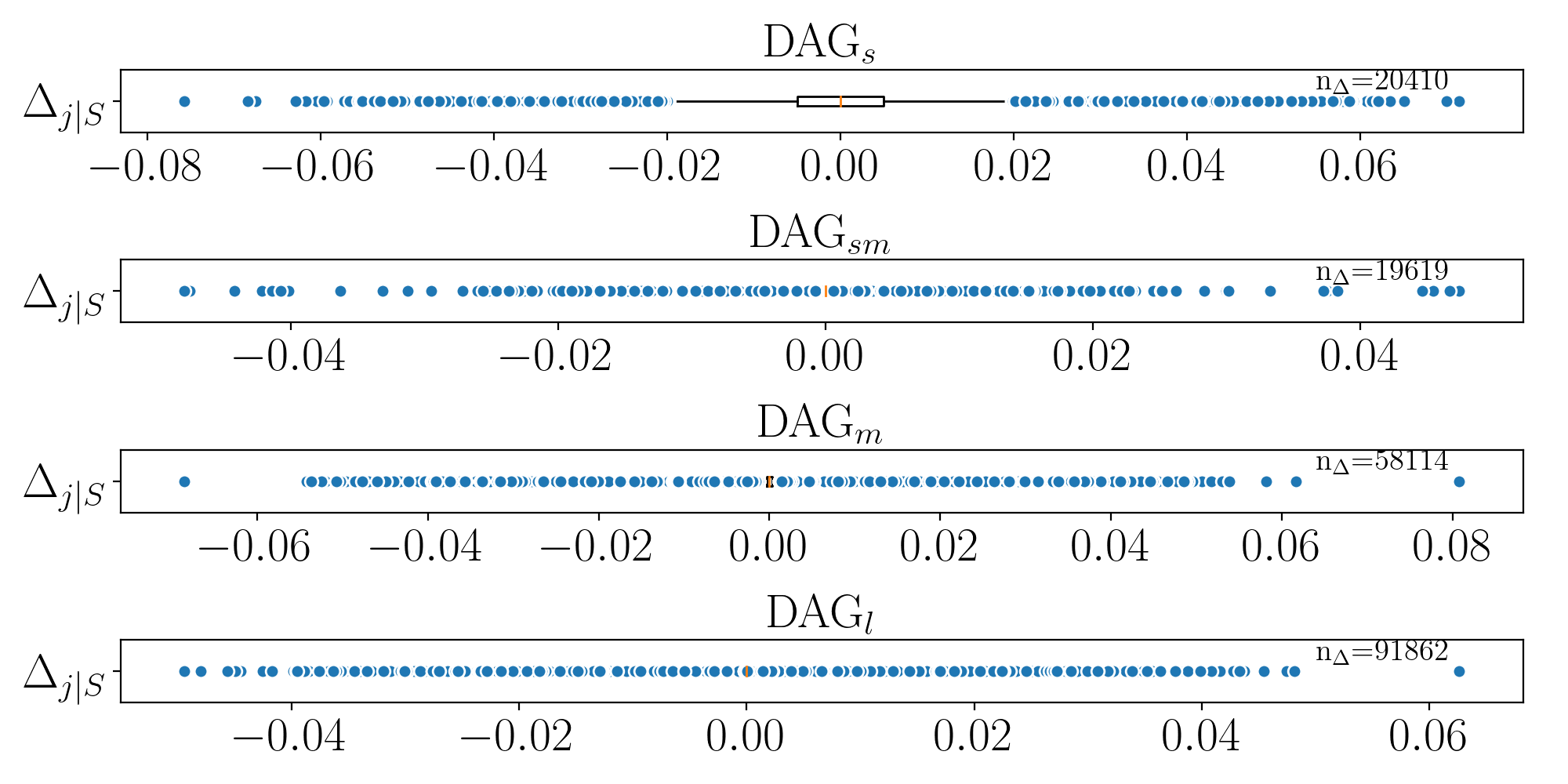}
            \caption{Boxplots showing the distribution of $\Delta_{j|S}$ for the skipped surplus evaluations.}
        \end{subfigure}%
    \caption{Results on the estimation quality for $d$-SAGE based on each DAG with average degree four and the LM. Based on five ($d$-)SAGE estimates.}
        \label{fig:sage-app-lm4}
\end{figure*}



\begin{figure*}[ht]
    \centering
        \begin{subfigure}{0.5\textwidth}
            \centering
                \begin{subfigure}{0.5\textwidth}
                    \centering
                        \includegraphics[width=0.99\linewidth]{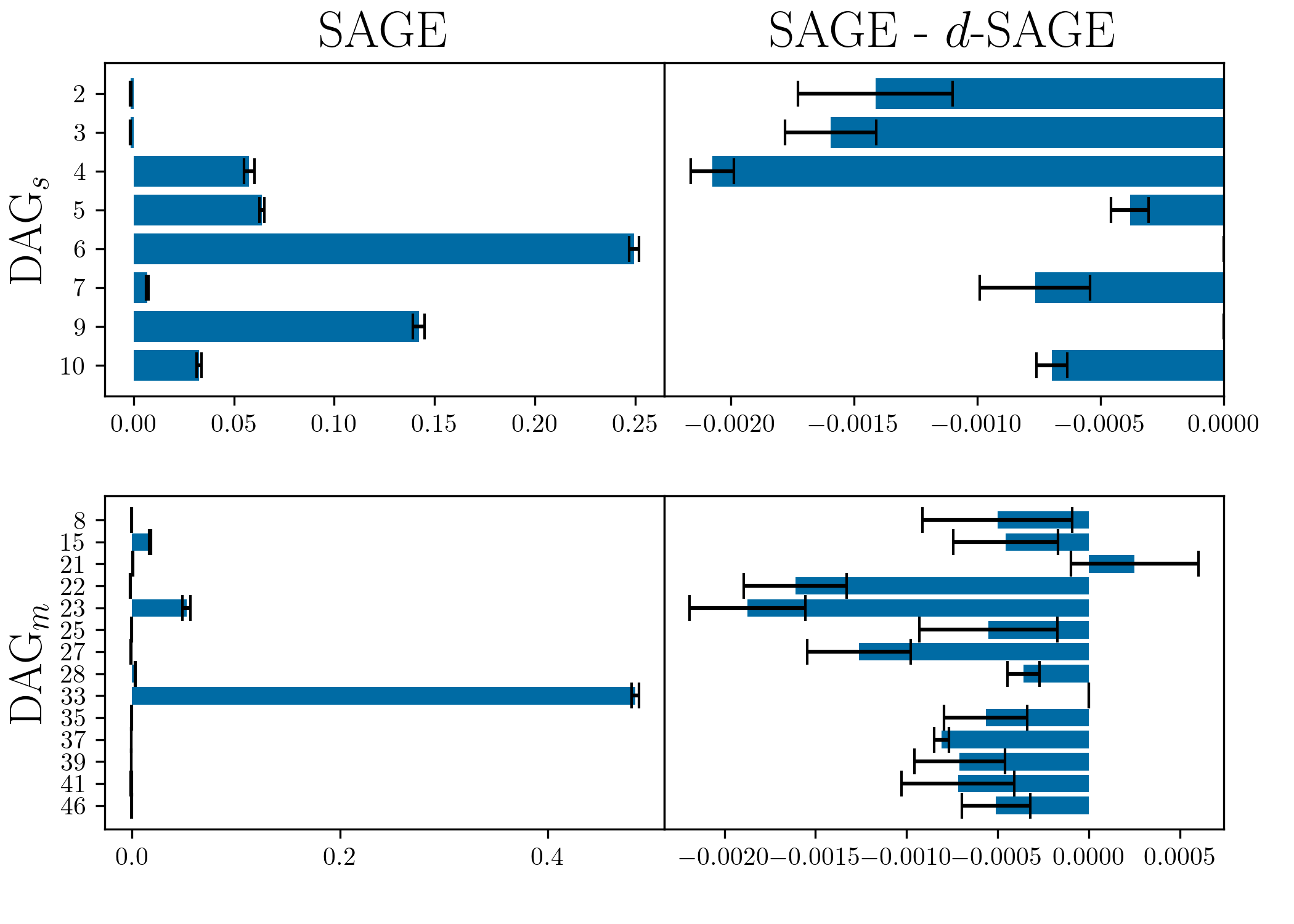}
                \end{subfigure}%
            \hfill
                \begin{subfigure}{0.5\textwidth}
                    \centering
                        \includegraphics[width=0.99\linewidth]{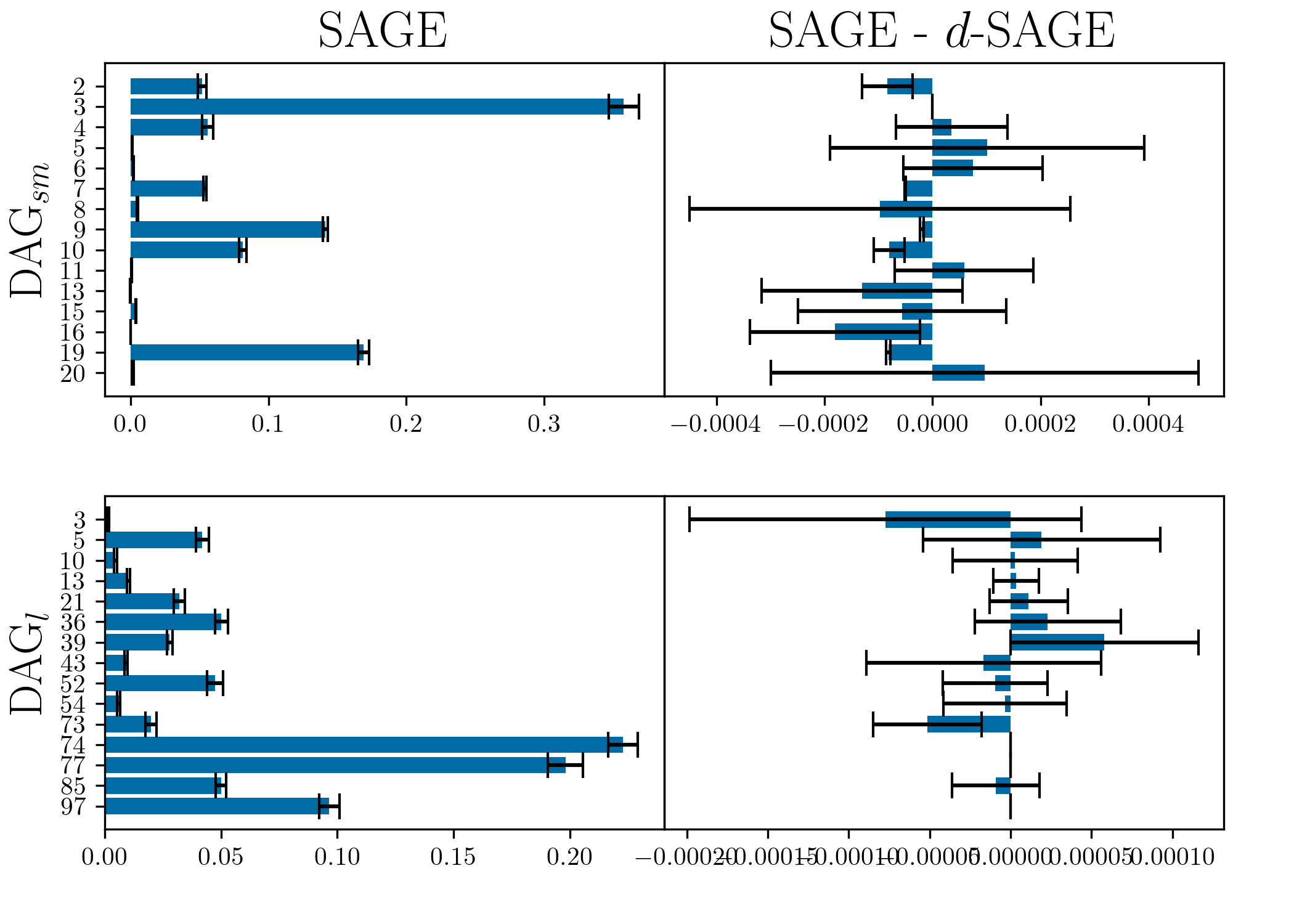}
                \end{subfigure}%
            \caption{SAGE values and difference between SAGE and $d$-SAGE for the fifteen (all for DAG$_s$) largest values for optimal models.}
        \end{subfigure}%
    \hfill
        \begin{subfigure}{0.5\textwidth}
            \centering
                \includegraphics[width=0.99\linewidth]{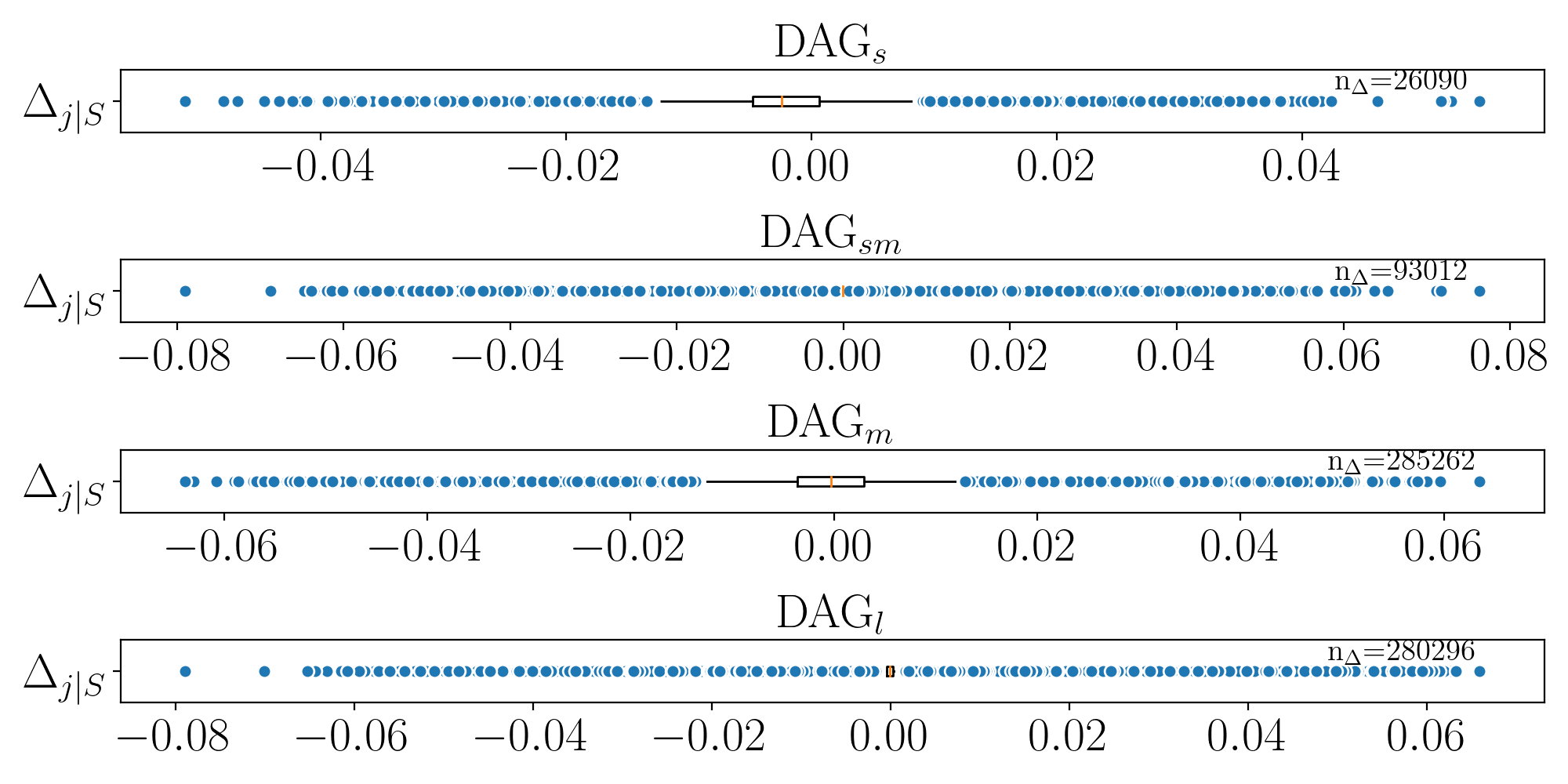}
            \caption{Boxplots showing the distribution of $\Delta_{j|S}$ for the skipped surplus evaluations.}     
        \end{subfigure}%
    \caption{Results on the estimation quality for $d$-SAGE based on each DAGs with average degree two and the RF. Based on five ($d$-)SAGE estimates.}
        \label{fig:sage-app-rf2}  
\end{figure*}



\begin{figure*}[ht]
    \centering
        \begin{subfigure}{0.5\textwidth}
            \centering
                \begin{subfigure}{0.5\textwidth}
                    \centering
                        \includegraphics[width=0.99\linewidth]{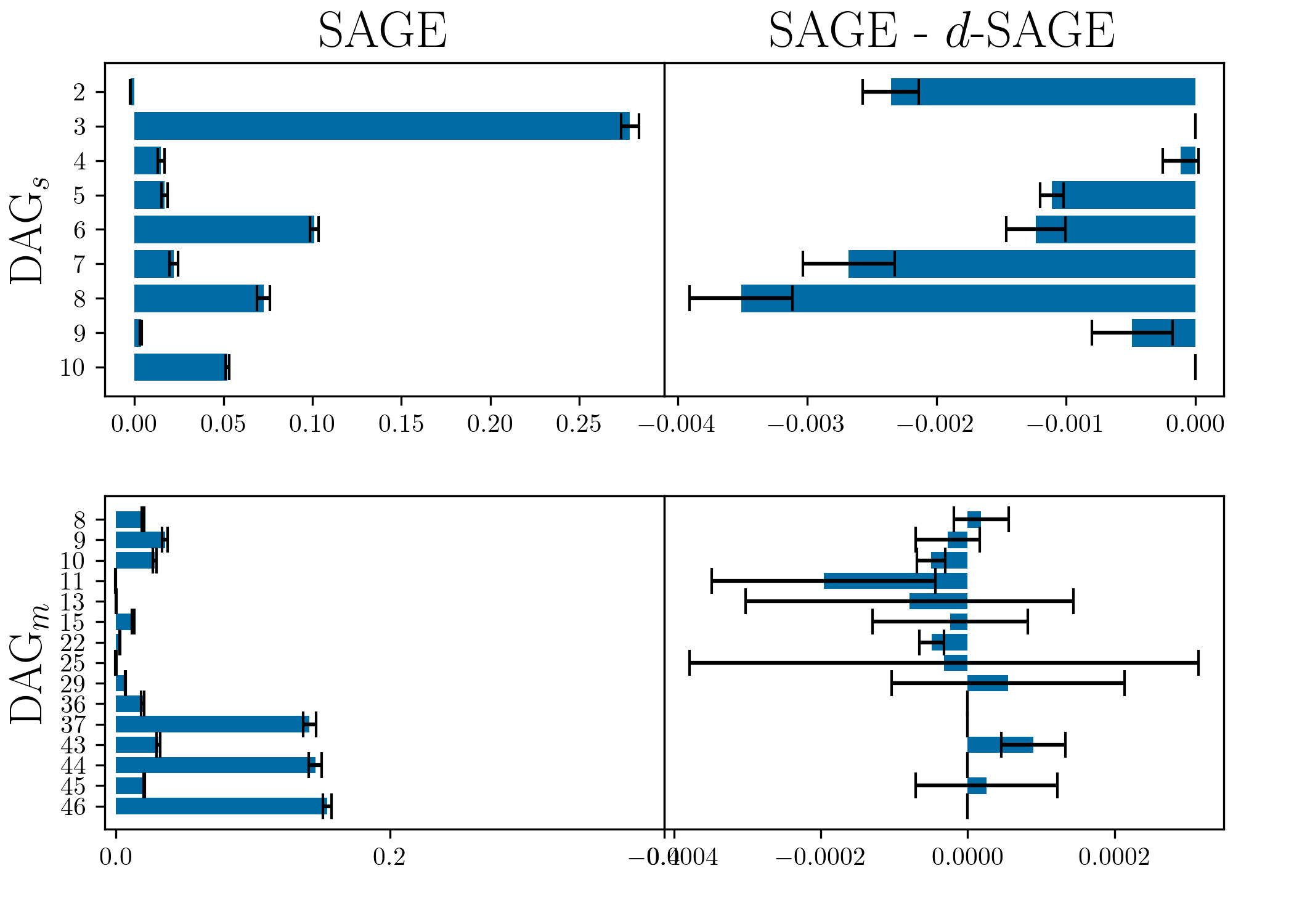}
                \end{subfigure}%
            \hfill
                \begin{subfigure}{0.5\textwidth}
                    \centering
                        \includegraphics[width=0.99\linewidth]{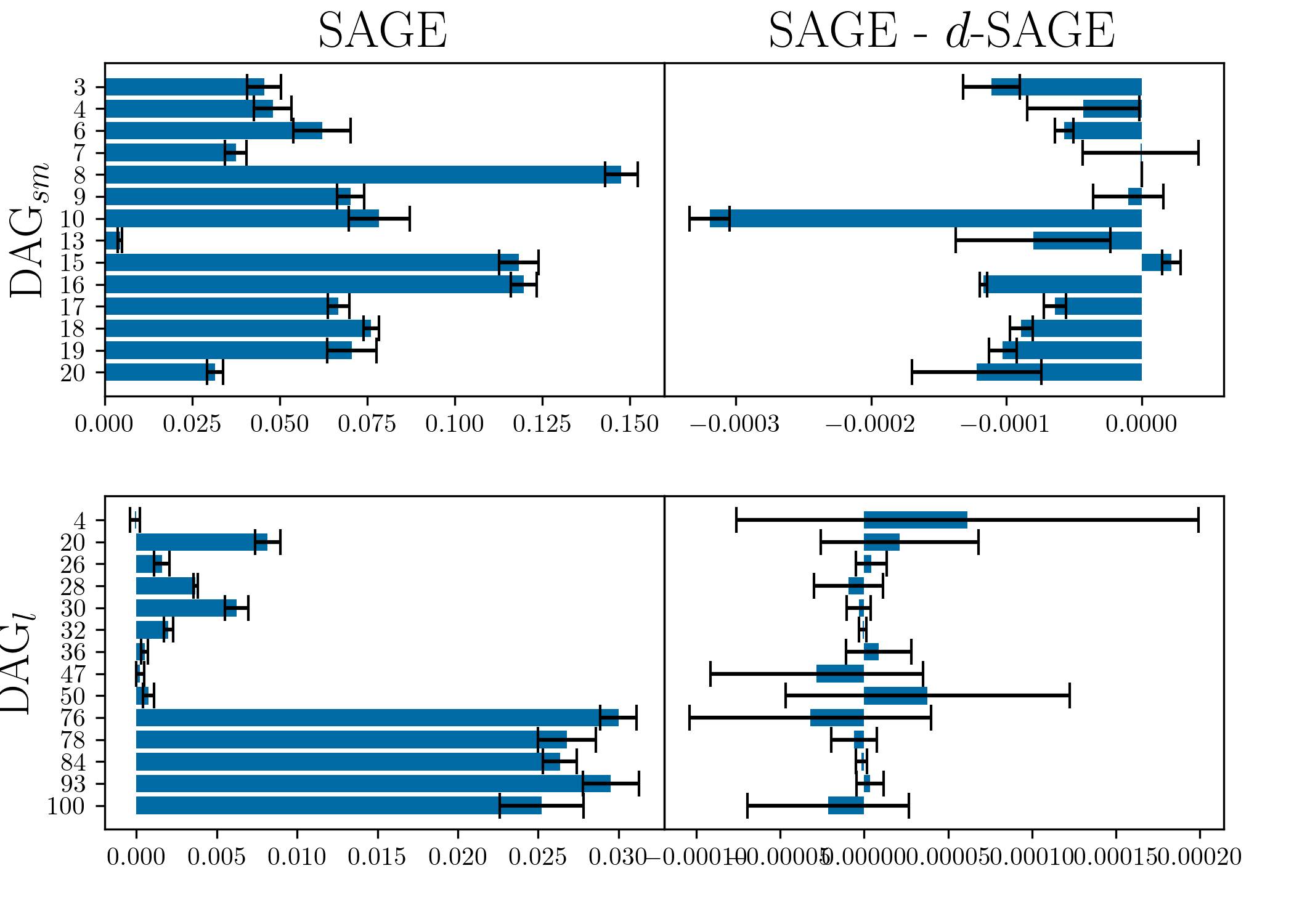}
                \end{subfigure}%
            \caption{SAGE values and difference between SAGE and $d$-SAGE for the fifteen (all for DAG$_s$) largest values for optimal models.}
        \end{subfigure}%
    \hfill
        \begin{subfigure}{0.5\textwidth}
            \centering
                \includegraphics[width=0.99\linewidth]{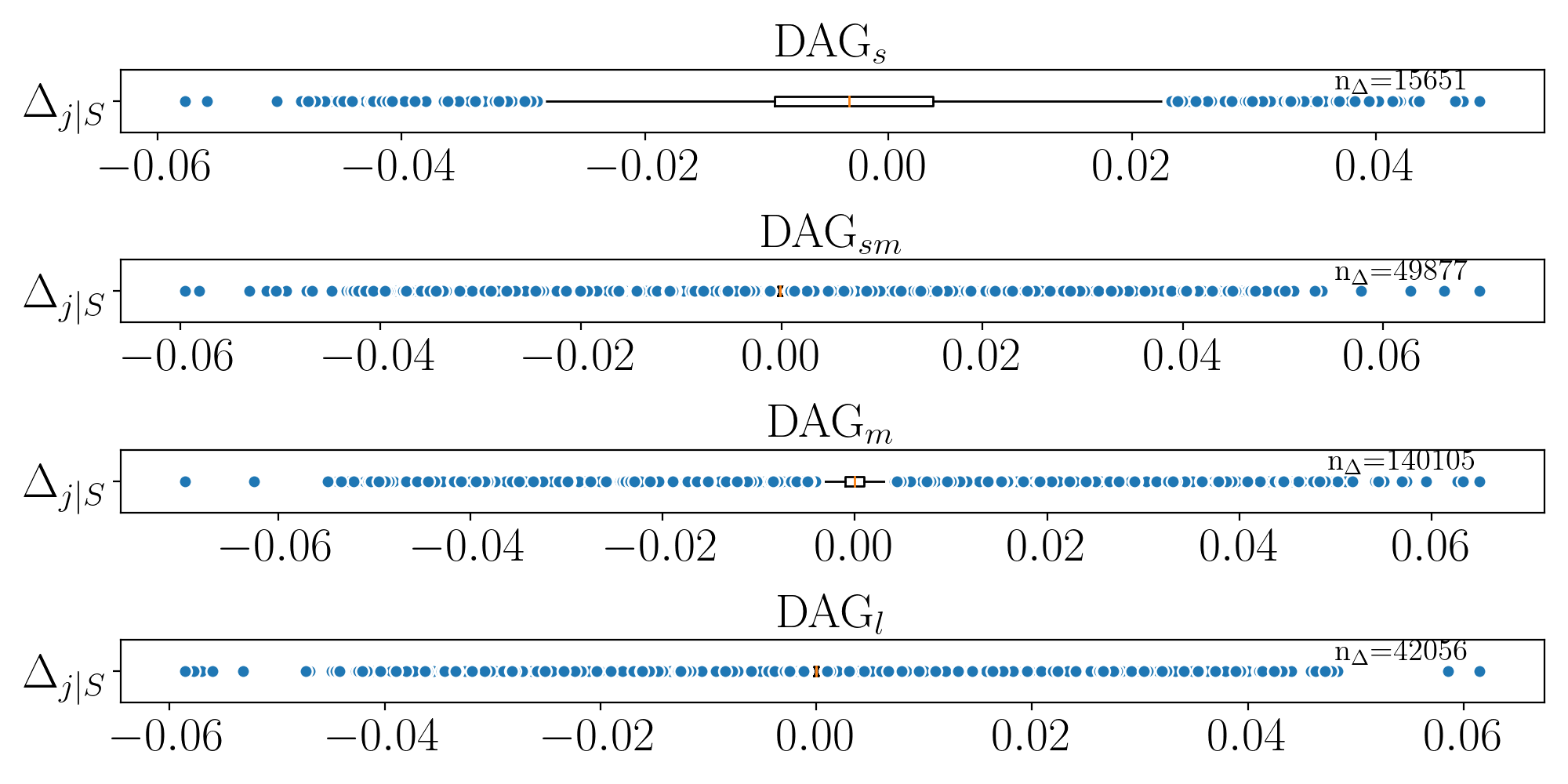}
            \caption{Boxplots showing the distribution of $\Delta_{j|S}$ for the skipped surplus evaluations.}     
        \end{subfigure}%
    \caption{Results on the estimation quality for $d$-SAGE based on each DAG with average degree three and the RF. Based on five ($d$-)SAGE estimates.}
        \label{fig:sage-app-rf3}  
\end{figure*}


\begin{figure*}[htb!]
    \centering
        \begin{subfigure}{0.5\textwidth}
            \centering
                \begin{subfigure}{0.5\textwidth}
                    \centering
                        \includegraphics[width=0.99\linewidth]{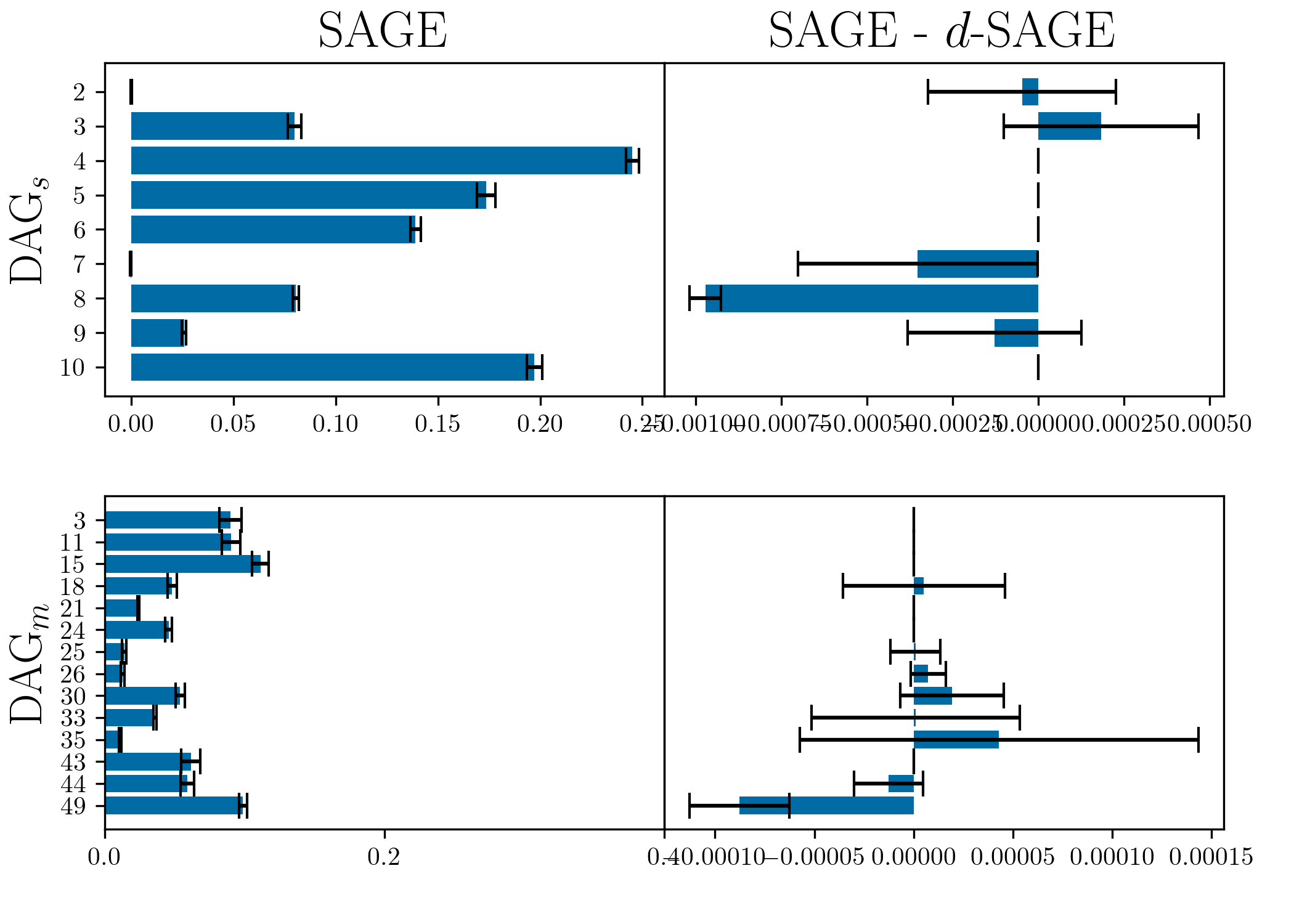}
                \end{subfigure}%
            \hfill
                \begin{subfigure}{0.5\textwidth}
                    \centering
                        \includegraphics[width=0.99\linewidth]{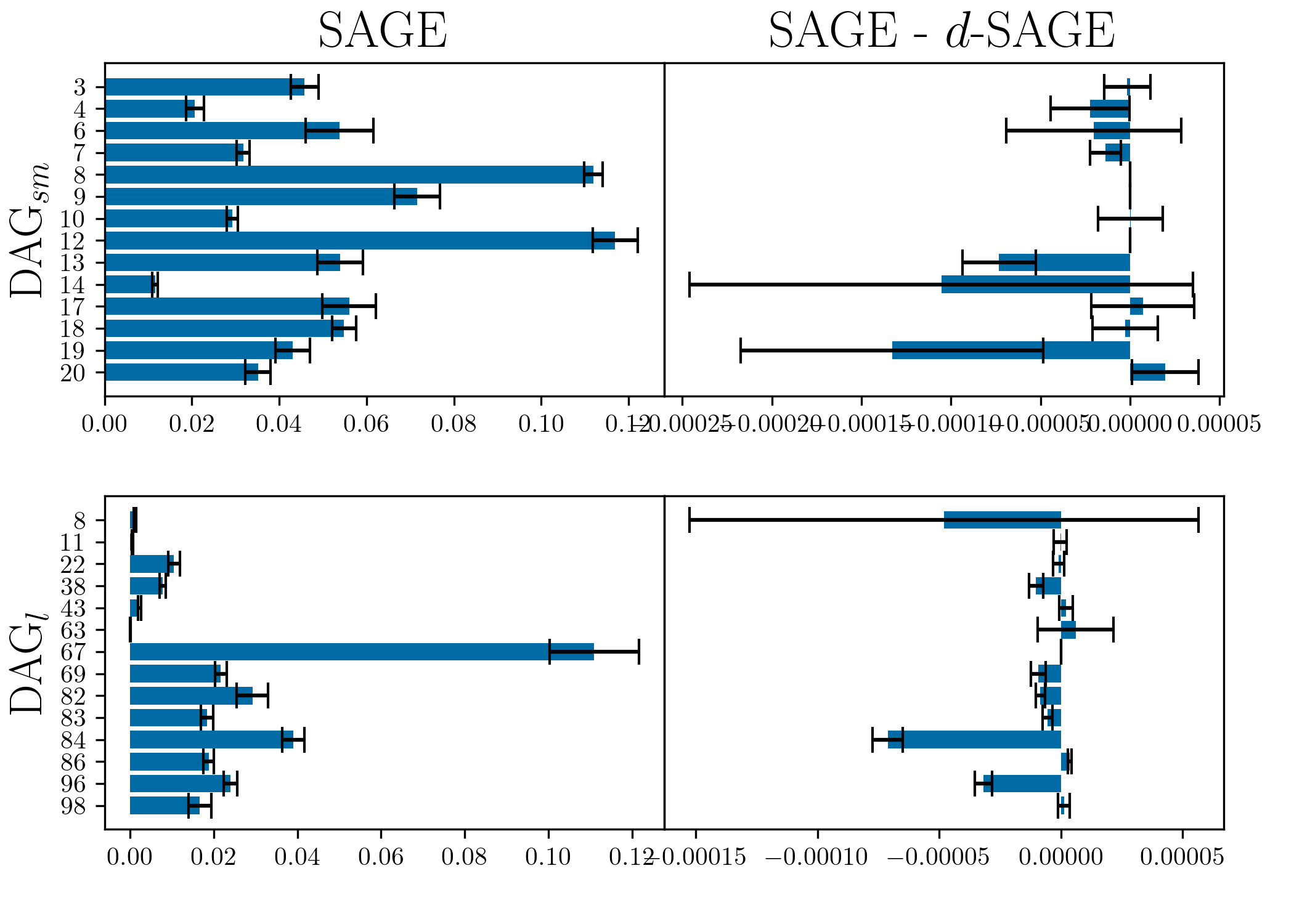}
                \end{subfigure}%
            \caption{SAGE values and difference between SAGE and $d$-SAGE for the fifteen (all for DAG$_s$) largest values for optimal models.}
        \end{subfigure}%
    \hfill
        \begin{subfigure}{0.5\textwidth}
            \centering
                \includegraphics[width=0.99\linewidth]{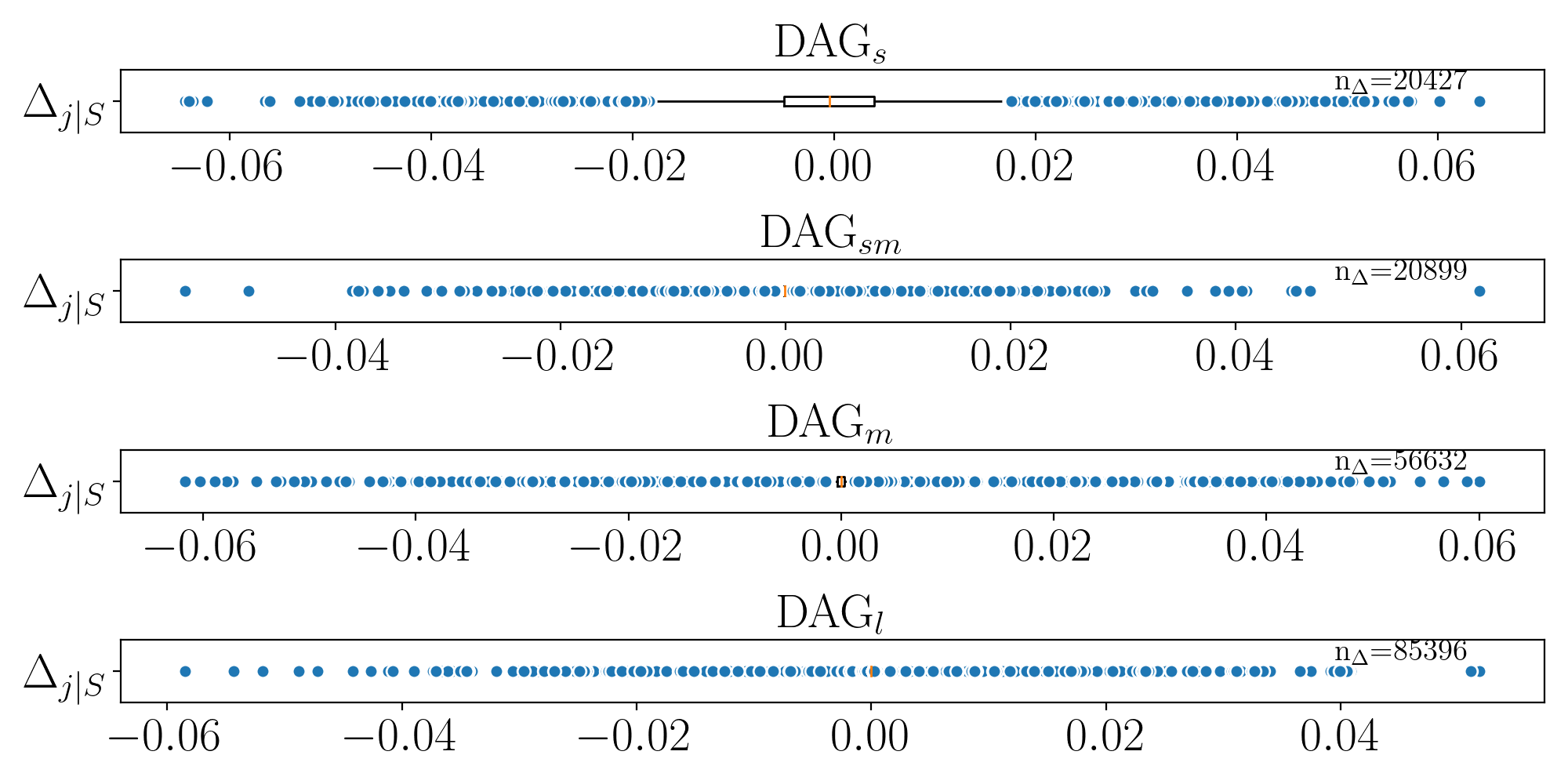}
            \caption{Boxplots showing the distribution of $\Delta_{j|S}$ for the skipped surplus evaluations.}     
        \end{subfigure}%
    \caption{Results on the estimation quality for $d$-SAGE based on each DAG with average degree four and the RF. Based on five ($d$-)SAGE estimates.}
        \label{fig:sage-app-rf4}  
\end{figure*}

\vfill

\clearpage

\section{CONVERGENCE PLOTS \label{appendix-convergence}}

\begin{figure*}[htb]
    \centering
        \includegraphics[scale=0.65]{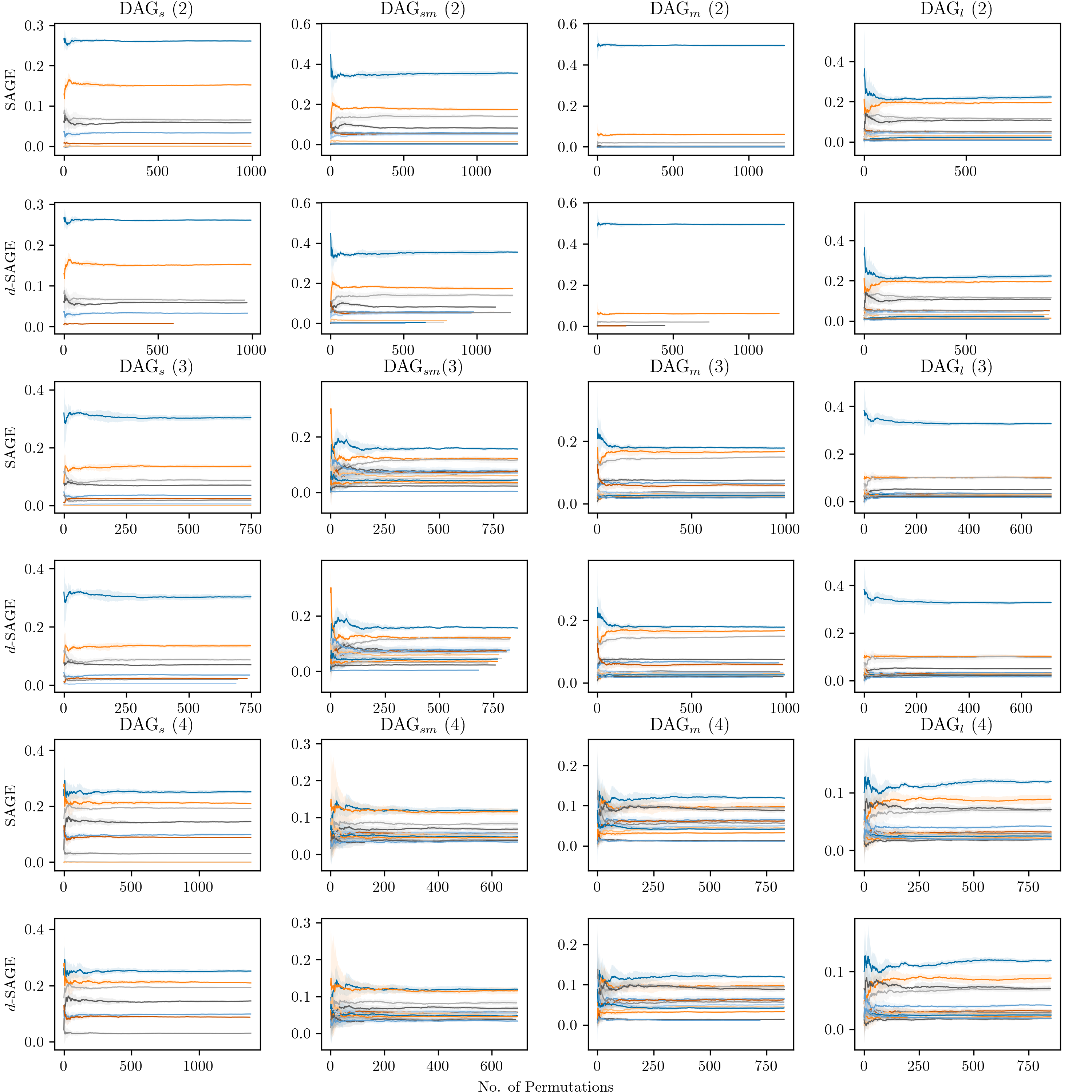}
    \caption{Convergence of largest fifteen SAGE and $d$-SAGE values for optimal models (LM) for every DAG (average adjacency degree). Each colour represents the same feature in SAGE and $d$-SAGE plots for a given graph (if present in both). Based on five ($d$-)SAGE estimates. Legend omitted for readability.}
        \label{fig:conv_lm}
\end{figure*}

\clearpage

\begin{figure*}[htb]
    \centering
        \includegraphics[scale=0.65]{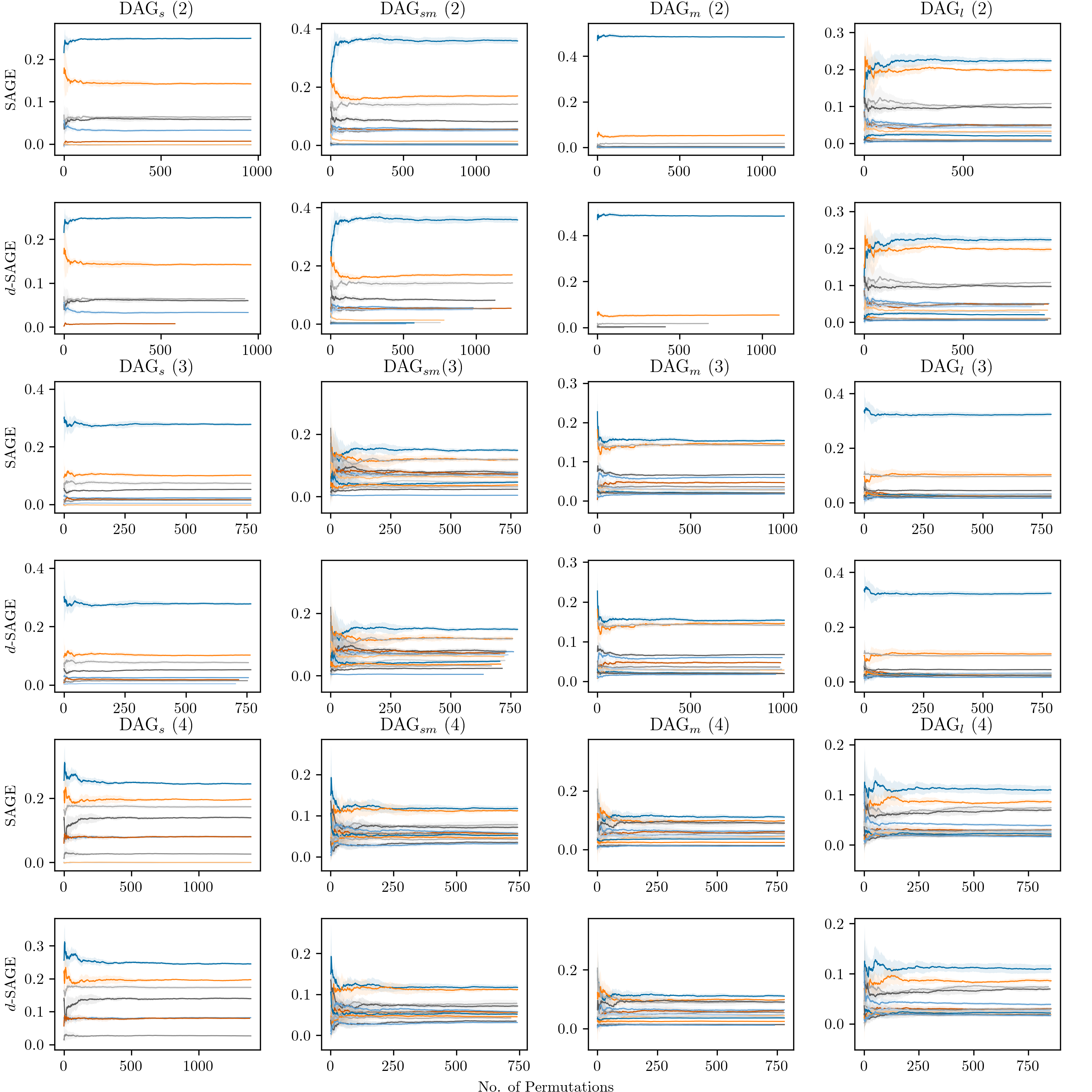}
    \caption{Convergence of largest fifteen SAGE and $d$-SAGE values for random forest models (RF) for every DAG (average adjacency degree). Each colour represents the same feature in SAGE and $d$-SAGE plots for a given graph (if present in both). Based on five ($d$-)SAGE estimates. Legend omitted for readability.}
        \label{fig:conv_rf}
\end{figure*}

\clearpage

\begin{figure*}[htb]
    \centering
        \includegraphics[scale=0.65]{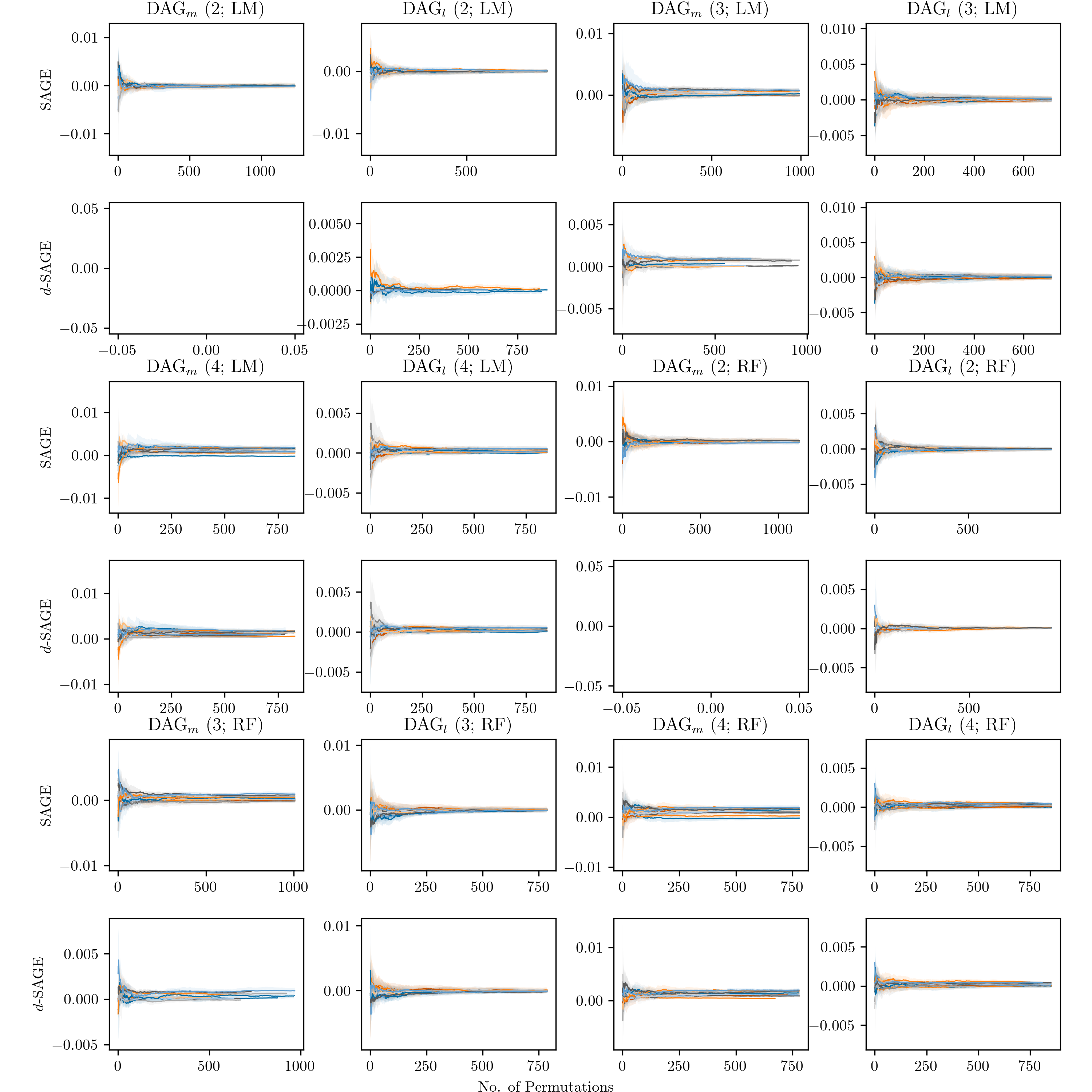}
    \caption{Convergence of bottom fifteen SAGE and $d$-SAGE values for optimal models (LM) and random forest models (RF) for DAG$_m$ and DAG$_l$ (average adjacency degree). Each colour represents the same feature in SAGE and $d$-SAGE plots for a given graph (if present in both). Based on five ($d$-)SAGE estimates. Legend omitted for readability.}
        \label{fig:conv_bottom}
\end{figure*}

\clearpage

\subsection{Convergence of SAGE Values}
The approximation algorithm is designed such that convergence for all values is required to stop. Hence, some values are converged but still computed. However, the benefit of $d$-SAGE depends on the share of CIs and not the number of permutations required for convergence, and hence, even a fewer number of permutations would lead to a similar speedup. Missing lines in the convergence plots belong to conditionally independent features (given every sampled coalition), which highlights the ability of ($d$-)SAGE for post-hoc feature selection. An example of faster converging $d$-SAGE values is displayed by the comparison of SAGE and $d$-SAGE for DAG$_m$(2) in Figure \ref{fig:conv_lm}, where the small values (slightly above zero) converge faster for $d$-SAGE. 

\section{PARTIAL CORRELATION TESTS vs. \texorpdfstring{$d$}{d}-SEPARATION QUERIES} \label{appendix-parr_corr}

To highlight the benefit of CSL over statistical independence tests, we compared the runtime of linear time $d$-separation queries (in graphs inferred by TABU) from the NetworkX package for Python \citep{hagberg2008networkx} to that of partial correlation tests for linear Gaussian data from the Pingouin package \citep{vallat2018ping}. Results are based on $100$ permutations, which amounts to the number of features ($9$, $19$, $49$ and $99$ for the different graphs) times $100$ as number of separate tests. Table \ref{fig:ci_v_dsep} clearly shows that partial correlation tests are typically more accurate at the cost of much higher runtime in comparison to $d$-separation queries (+ graph learning).

\begin{table}[ht]
\caption{Partial correlation tests v $d$-separation queries based on $n=10,000$ and $100$ permutations as in ($d$-)SAGE evaluation; Graph learning based on TABU; ACC = Accuracy.} \label{fig:ci_v_dsep} \label{sample-table}
\begin{center}
\begin{tabular}{l c c c c c c}
\textbf{DATA}  & \textbf{TIME ($d$-separation)} & \textbf{TIME (TABU)} & \textbf{TIME (CIs)} & \textbf{ACC ($d$-separation)} & \textbf{ACC (CIs)} \\
\hline \\
DAG$_s$ (2)         & 0.13s & 0.06s & 46.82s & 1.000 & 1.000 \\
DAG$_{sm}$ (2)      & 0.39s & 0.22s & 166.75s & 0.996 & 0.999 \\
DAG$_m$ (2)         & 1.95s & 1.11s & 1058.80s & 1.000 & 1.000 \\
DAG$_l$ (2)         & 15.48s & 12.02s & 4344.81s & 0.863 & 0.934 \\
DAG$_s$ (3)         & 0.13s & 0.18s & 47.15s & 1.0 & 1.0 \\
DAG$_{sm}$ (3)      & 0.41s & 0.71s & 166.28s & 0.996 & 0.992 \\
DAG$_m$ (3)         & 2.12s & 2.51s & 1089.00s & 0.908 & 0.983 \\
DAG$_l$ (3)         & 16.85s & 18.22s & 4299.47s & 0.857 & 0.941 \\
DAG$_s$ (4)         & 0.14s & 0.09s & 47.18s & 1.0 & 0.998 \\
DAG$_{sm}$ (4)      & 0.42s & 1.37s & 163.48s & 1.0 & 0.988 \\
DAG$_m$ (4)         & 2.33s & 5.65s & 1093.50s & 0.845 & 0.940 \\
DAG$_l$ (4)         & 20.74s & 39.86s & 4312.16s & 0.902 & 0.916 \\

\end{tabular}
\end{center}
\end{table}

\vfill

\end{document}